\documentclass[10pt,twocolumn,letterpaper]{article}

\usepackage{cvpr}              

\usepackage{graphicx}
\usepackage{amsmath}
\usepackage{algorithmic}
\usepackage{algorithm}
\usepackage{amssymb}
\usepackage{booktabs}
\usepackage{color, colortbl}
\usepackage{enumitem}

\usepackage{listings}
\usepackage[autolanguage]{numprint} 
\definecolor{commentcolor}{RGB}{110,154,155}   

\usepackage[pagebackref,breaklinks,colorlinks]{hyperref}

\usepackage[capitalize]{cleveref}
\crefname{section}{Sec.}{Secs.}
\Crefname{section}{Section}{Sections}
\Crefname{table}{Table}{Tables}
\crefname{table}{Tab.}{Tabs.}

\def\cvprPaperID{9180} 
\def\confName{CVPR}
\def\confYear{2022}

\begin{document}
\title{Unseen Classes at a Later Time? No Problem}

\author{Hari Chandana Kuchibhotla$^{\ddag}$\thanks{equal contribution} , Sumitra S Malagi$^{\ddag}$\footnotemark[1] , Shivam Chandhok $^{\diamond}$ $^{\ddag}$, Vineeth N Balasubramanian$^{\ddag}$\\
\normalsize $^{\ddag}$ Indian Institute of Technology Hyderabad, India  $^{\diamond}$ INRIA, Universite Grenoble Alpes  \\
{\tt\small \{ai20resch11006,cs20mtech14006,vineethnb\}@iith.ac.in,chandhokshivam@gmail.com}
}

\maketitle

\begin{abstract}
Recent progress towards learning from limited supervision has encouraged efforts towards designing models that can recognize novel classes at test time (generalized zero-shot learning or GZSL). GZSL approaches assume knowledge of all classes, with or without labeled data, beforehand. However, practical scenarios demand models that are adaptable and can handle dynamic addition of new seen and unseen classes on the fly (i.e continual generalized zero-shot learning or CGZSL). One solution is to sequentially retrain and reuse conventional GZSL methods, however, such an approach suffers from catastrophic forgetting leading to suboptimal generalization performance.
A few recent efforts towards tackling CGZSL have been limited by difference in settings, practicality, data splits and protocols followed -- inhibiting fair comparison and a clear direction forward. 
Motivated from these observations, in this work, we firstly consolidate the different CGZSL setting variants and propose a new Online-CGZSL setting which is more practical and flexible. Secondly, we introduce a unified feature-generative framework for CGZSL that leverages bi-directional incremental alignment to dynamically adapt to addition of new classes, with or without labeled data, that arrive over time in any of these CGZSL settings. Our comprehensive experiments and analysis on five benchmark datasets and comparison with baselines show that our approach consistently outperforms existing methods, especially on the more practical Online setting. 
\end{abstract}

\begin{figure}
    \centering
    \includegraphics[width=0.45\textwidth]{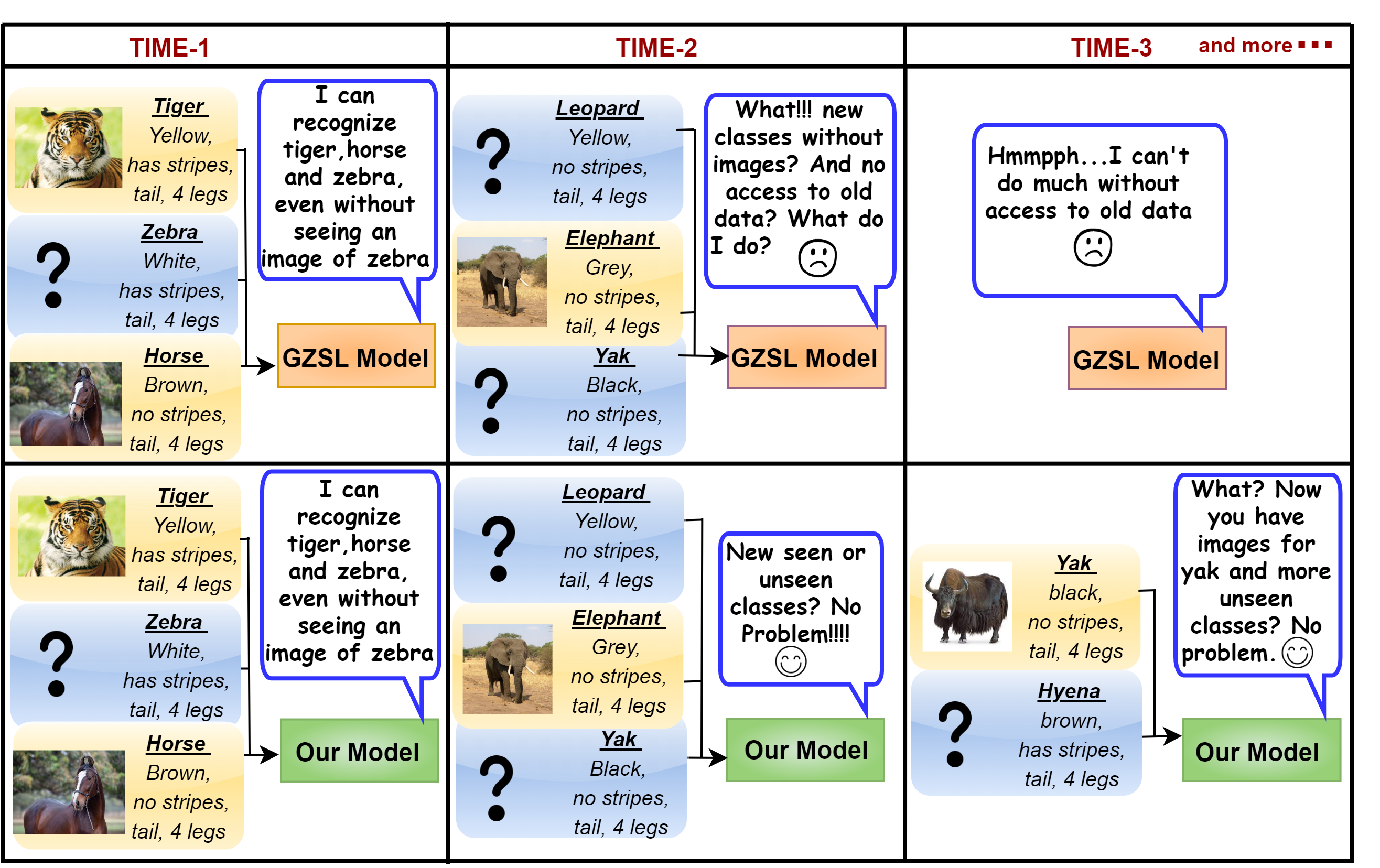}
    \vspace{-2pt}
    \caption{\footnotesize \textbf{Illustration of proposed setting.} Applicability of a Generalized Zero-shot Learning (GZSL) model (row 1) and our proposed model (row 2) in a real-world scenario on dynamic addition of new seen (depicted in yellow) and unseen (depicted in blue) categories. The GZSL setting does not allow dynamic addition of newer classes, seen or unseen, that are added over time. Our proposed CGZSL setting is more flexible and can tackle dynamic changes in the initial pool of seen and unseen categories, enhancing model scalability and applicability in practical settings.}
    \vspace{-10pt}
    \label{introfig}
\end{figure}

\vspace{-8pt}
\section{Introduction}
\vspace{-4pt}
Deep Neural Networks (DNNs) have shown great promise as predictive models and are increasingly being used for various computer vision applications.
However, their reliance on large-scale labeled datasets limit their use in practical scenarios encountered in the real-world.
The occurrence of objects in the real world inherently follows long-tailed distributions\cite{openlongtailrecognition,zhang2021deep}, implying that visual data sampled from the real world may not be readily available for all categories of interest at the same time. Thus, it is required for models to have the ability to generalize and recognize novel objects semantically similar to the ones encountered during training even though visual data for these novel classes is not seen by the model. Existing efforts aim to tackle this problem, by designing generalized zero-shot learning (GZSL) models that are equipped with the ability to generalize to unseen classes at test time.

Another important aspect of object occurrence in the real world is the gradual addition of object categories with time. This can be attributed to the discovery of new objects or the continuous nature of data collection process owing to which a previously rare object may have abundantly available samples at a later stage. However, existing GZSL approaches are not designed to tackle dynamic addition of classes in the initial pool of seen and unseen categories, limiting their scalability and applicability in challenging practical settings.
Fig. \ref{introfig} illustrates the shortcomings of a GZSL model in such practical settings where classes arrive over time. Evidently, the GZSL setting is limiting, and is unable to adapt to dynamic changes to the initial pool of categories brought by the gradual addition of new seen and unseen classes over time. This can be attributed to the fact that since the previous data is no longer available, the GZSL model tends to \textit{catastrophically forget} knowledge pertaining to previous tasks when sequentially trained and reused over time. This necessitates concerted effort toward carefully designing problem settings that resemble the occurence of objects in the real world, and building models that can adapt over time and seamlessly tackle such challenges.

Very recently, sporadic efforts \cite{cln,ghosh,iisc} have been made towards designing models that can dynamically adapt and generalize on addition of new seen and unseen classes. The aforementioned works term this setting as \textit{continual generalized zero-shot learning} (CGZSL). However, these efforts are nascent, and vary in the definition of the problem setting, practicality, data splits and protocols followed -- thus inhibiting fair comparison and a clear path forward. 
To address this issue and motivated by the need for progress in this direction, in this work, we firstly consolidate the different CGZSL settings tackled in these recent efforts, and clearly segregate existing methods and settings according to the challenges they tackle. We propose a more flexible and realistic \emph{Online-CGZSL} setting which more closely resembles scenarios encountered in the real world. In addition, we propose a unified framework that employs a bi-directional incremental alignment-based replay strategy to seamlessly adapt and generalize to new seen and unseen classes that arrive over time. Our replay strategy is based on a feature-generative architecture, and hence does not require storing samples from previous tasks. We also use a static architecture for incremental learning (as against using a model-growing one) in order to facilitate scalability and efficiency. 
In summary, the key contributions of our work are below:
\vspace{-4pt}
\begin{itemize}[leftmargin=*]
\setlength\itemsep{-0.09em}
\item We identify the different challenges of the relatively new CGZSL setting, and consolidate its different variants based on the challenges they tackle and restrictions they impose. We hope that this will enable fair comparison among such approaches and further progress in the field.
\item  We establish a practical, but more challenging, \textit{Online-CGZSL} setting which more closely resembles real-world scenarios encountered in practice. 
\item We propose a novel feature-generative framework to address the different CGZSL setting variants, which avoids catastrophic forgetting through bi-directional incremental alignment thereby allowing forward semantic knowledge transfer from previous tasks and enabling generalization.  

\item We perform extensive experiments and analysis on three different CGZSL settings on well-known benchmark datasets: AWA1, AWA2, Attribute PASCAL and Yahoo (aPY), Caltech-UCSD-Birds (CUB) and SUN, demonstrating the promise of our approach. We observe that our model consistently improves over baselines and existing approaches, especially on the more challenging \emph{Online} setting.
\end{itemize}
\begin{figure*}[t]
    \centering
    \includegraphics[width=1.00\textwidth]{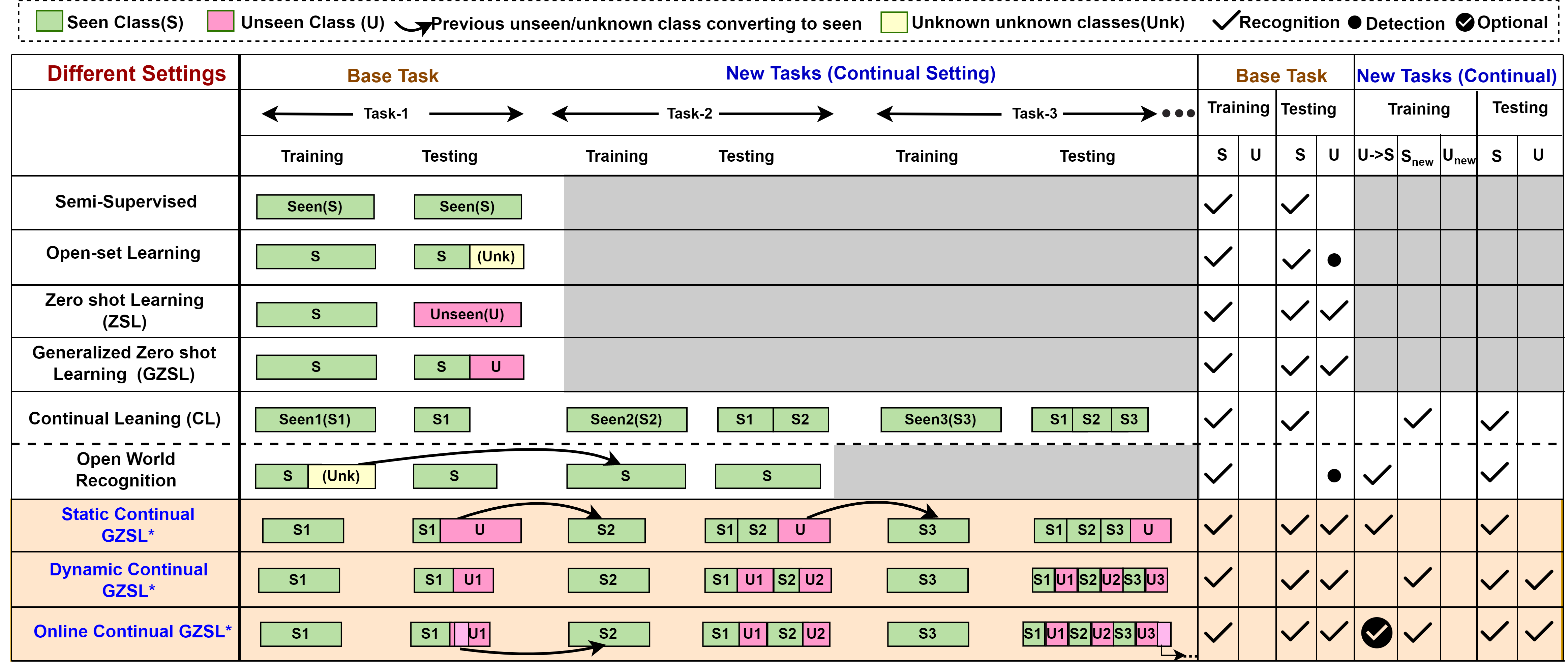}
    \vspace{-16pt}
    \caption{\footnotesize \textbf{Comparison of proposed settings with other known settings.} In ZSL, model is trained on seen classes and evaluated on unseen classes. In GZSL, the model is tested on both seen and unseen classes. CL models are trained on classes that arrive sequentially, but do not have unseen classes either during training or testing. Continual GZSL (CGZSL) settings (highlighted with a *) are proposed in this work. In static- CGZSL, classes that arrive in future are considered unseen. In Dynamic-CGZSL, each task has disjoint set of seen and unseen classes. Online-CGZSL allows for conversion of previously unseen classes to seen (based on availability of data) in addition to handling new seen and unseen classes at each task.}
    \vspace{-6pt}
    \label{fig:settings}
\end{figure*}
\section{Related Work}
\vspace{-4pt}
\noindent \textbf{Generalized Zero-Shot Learning (GZSL).}
Existing GZSL approaches can be broadly classified into \textit{embedding-based methods} and \textit{generative methods}. Traditional embedding-based approaches \cite{zeroshotvis,devise,simplezsl,deepembed} aim to project visual and/or semantic features onto a common embedding space and use a nearest neighbor based classifier to classify visual samples. 
On the other hand, a relatively recent and more effective set of approaches \cite{lisgan,shivam,cvae,zerovaegan,fclswgan} propose to use generative models to synthesize unseen visual features, thus converting the GZSL problem into a supervised classification problem.
Although, the aforementioned approaches aim to generalize to novel unseen classes at test-time, they are not designed to tackle dynamic addition of categories (seen or unseen) over time.

\noindent \textbf{Continual Learning (CL).}
Existing approaches that aim to learn continually and tackle catastrophic forgetting \cite{catasinfer} can broadly be categorized into: parameter isolation methods \cite{packnet}, regularization-based methods \cite{ewc,sdc,lwf}, model-growing architectures\cite{pnn}, and rehearsal-based methods \cite{agem}. Parameter isolation methods \cite{packnet} aim to identify the parameters important for a task while regularization-based methods \cite{ewc,sdc,lwf} constrain parameters to avoid deviation of weights important to previous tasks. On the other hand,  model growing architectures \cite{clbayes} dynamically increase model capacity and rehearsal-based methods store or generate images of previous tasks to avoid forgetting \cite{genclassinc}.  In addition, various approaches use knowledge distillation technique\cite{genclassinc,icarl} to transfer knowledge from previous tasks to current task. However, these approaches cannot generalize to unseen classes for which visual samples have not been encountered by the model during training.

\noindent \textbf{Continual Generalized Zero-Shot Learning (CGZSL).} 
While CL aims to learn incrementally by \textit{transferring learned knowledge} to future tasks, GZSL aims to \textit{transfer semantic knowledge} from seen classes to recognize unseen classes. Considering the common objective of transferring learned knowledge, there have been a few recent efforts to unify these paradigms. Lifelong ZSL\cite{lzsl} marked the first attempt to address the problem of CGZSL, where a multihead architecture was used to accumulate knowledge while training from multiple datasets. However, the approach requires task level supervision in the form of task-ids at test time limiting its applicability in realistic scenarios.  On the other hand, \cite{cln} suggested a class normalization-based approach to solve the CGZSL problem and \cite{ghosh} learned a new VAE for every task. However, \cite{cln,bookworm,ghosh} specifically consider previously encountered tasks as seen and future tasks as unseen classes, and thus only tackle the \emph{static-CGZSL} setting which is restrictive in practice (Sec.\ref{cgzslwhat}). 
\cite{iisc,iisc2} recently formulated CGZSL as a problem where each task has its own set of seen and unseen classes and work in the \emph{dynamic-CGZSL} setting. 
Although this setting is an improvement over \emph{static-CGZSL} setting, it is still restrictive as it does not allow the dynamic conversion of unseen classes to seen as data becomes available over time.
Furthermore, learning a new VAE for every task \cite{ghosh} or storing exemplar samples of previous tasks \cite{iisc2} results in progressively increasing memory requirements, which is inefficient. On the other hand, \cite{iisc} employs a rehearsal based strategy but does not take into account the changing semantic structure of visual spaces due to addition of new categories over time.

\begin{figure*}[t]
    \centering
    \includegraphics[width=0.8\textwidth]{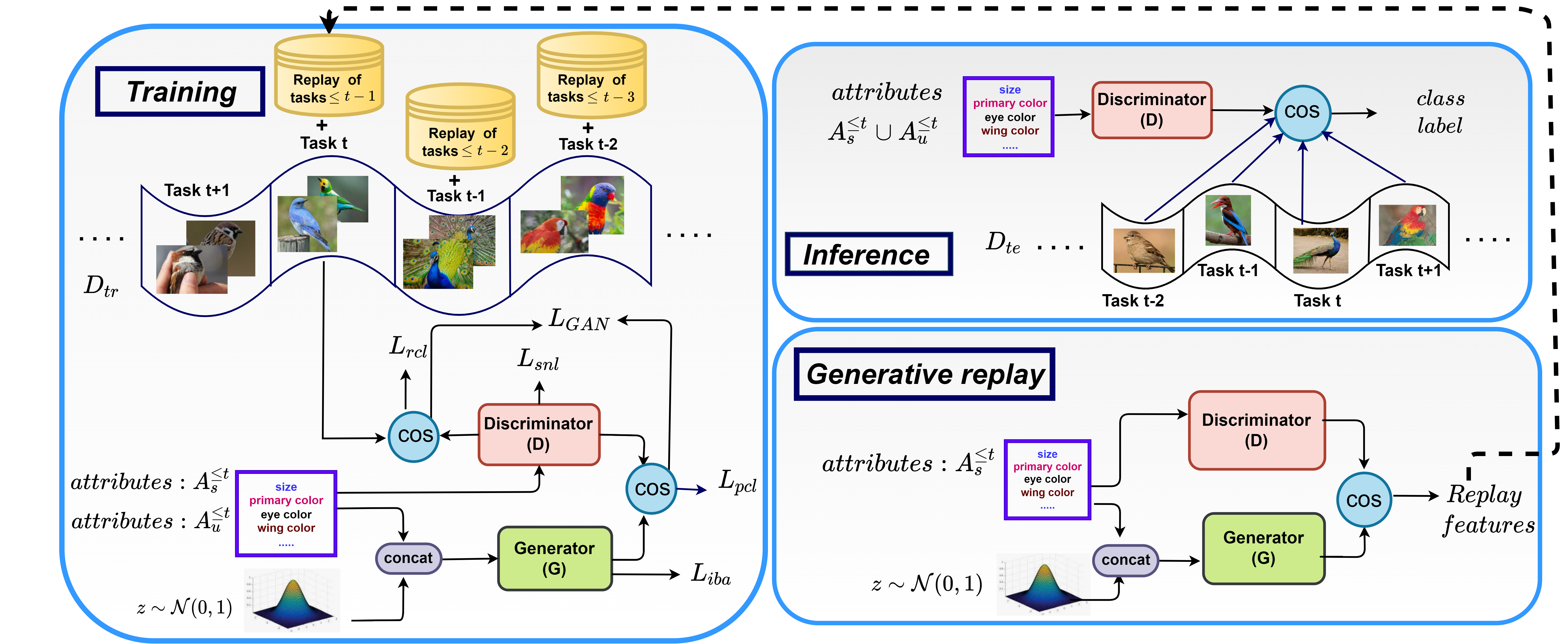}
    \vspace{-4pt}
    \caption{\footnotesize Our approach consists of three phases: (1) Training; (2) Generative replay; and (3) Inference. (1) During training, the discriminator $D$ is trained to map input attribute to the visual space, and generator $G$ is trained to generate pseudo-visual features. The GAN  is trained adversarially using similarity scores. $L_{rcl}$, $L_{pcl}$ and $L_{snl}$ loss terms aid in generating discriminative features for seen and unseen classes. $L_{sal}$ and $L_{nuclear}$ are used for incremental bi-directional alignement; (2) We test on three different settings at a given time step $t$, as described in Sec. \ref{cgzslwhat}. Cosine similarity is used to classify the target visual feature; (3) Trained $G$ is used to replay/generate features of seen classes encountered until task $t$.}
    \vspace{-9pt}
    \label{fig:my_label}
\end{figure*}

\vspace{-4pt}
\section {CGZSL: Settings and Formulation}
\label{cgzslwhat}
\vspace{-6pt}
In this section, we consolidate and provide a holistic overview of the various CGZSL settings tackled by recent efforts and discuss in detail the major differences in their formulations. We also describe our proposed \emph{Online-CGZSL} setting which is more flexible and resembles real world scenarios more closely. Fig. \ref{fig:settings} illustrates a pictorial representation of the various CGZSL settings and strives to correlate and appropriately position them w.r.t. other related limited-supervision and continual learning settings. We now describe each of the CGZSL setting variants below.

\vspace{-11pt}
\paragraph{Static-CGZSL.}
In this setting \cite{cln,bookworm,ghosh,azsl}, the dataset is divided into $T$ subsets, and the model encounters each of these subsets in an incremental fashion over time. The setting assumes all previously encountered tasks as seen and future tasks as consisting of unseen classes. Formally, for a given task $\mathcal{T}_t$ at a given time step $t$, the first $t$ subsets i.e data belonging to the current and previous tasks are considered as seen classes while the future tasks are considered unseen.

This setting differs from traditional GZSL in the fact that during evaluation of the $t^{th}$ task, previous training data is unavailable. Thus the model should be capable of retaining previously learned knowledge while adapting to the newly encountered seen classes. However,
\emph{static-CGZSL} presents a constrained setting which requires the total number of classes or tasks to be known beforehand (hence the name $static$).
Furthermore, the setting mandates that all tasks till current time step $t$ are considered as seen classes, with only the future tasks contain the unseen classes. Thus the dynamic addition of classes is restricted to conversion of an unseen class to seen after a particular task is encountered while learning continually. While it is reasonable to assume that visual features of unseen classes may become available in future, it may be non-viable to assume that novel seen or unseen categories, unknown at the beginning of training, will not be added in future.  
This fundamentally limits the notion of continual learning where a model should be adaptable to any number of tasks or addition of new classes. 

\vspace{-11pt}
\paragraph{Dynamic-CGZSL.}
Considering the limitations of \emph{static-CGZSL} setting, \cite{iisc,iisc2} proposed another setting where each task has an exclusive set of seen and unseen classes and the model can accommodate any number of tasks over time. We categorize this setting as dynamic-CGZSL, which is less restrictive than \emph{static-CGZSL} setting as it allows for addition of both seen and unseen classes in a continual manner. However, this setting imposes a constraint that previously unseen classes may never become seen in future and is unable to tackle the scenario where a rare class may have abundantly available samples at a later time stage. This is limiting in practice as owing to the continuous nature of data collection, it is plausible that some of the unseen classes may become seen  when  its  data  becomes  available over time.

\vspace{-11pt}
\paragraph{Online-CGZSL.}
In order to better align with the scenarios commonly encountered in the real-world, we introduce a new \emph{Online-CGZSL} setting which can handle a variety of dynamic changes in the pool of seen and unseen classes. Specifically, each task has a disjoint set of seen and unseen classes, and an arbitrary number of such tasks/categories can be dynamically incorporated on the fly. Importantly, this setting allows the conversion of previously unseen classes into seen if the corresponding visual features become available depending on changes in data availability in the future. Note that our setting is more flexible as it does not require the knowledge of entire pool of categories beforehand, impose any restrictions on the conversion of unseen class to seen class or require task level supervision at test-time. Next, we formally describe and formulate our proposed \emph{Online-CGZSL} problem setting.


\vspace{-11pt}
\paragraph{Problem Formulation.}
Let subscripts $s$ and $u$ denote seen and unseen classes respectively. Each task $t$ consists of train and test data. Let $A^t$ be the union of attributes of seen ($A_{s}^{t}$) and unseen ($A_{u}^{t}$) encountered so far. The training data $D_{tr}^{t}$ at task $t$ is given by $D_{tr}^{t}$ = \{( training visual samples of current task $t$ ($X_{s_{tr}}^{t}$), their class labels ( $Y_{s_{tr}}^{t}$), class attributes($A^t$))\}.  The test data at task $t$ be $D_{te}^{t}$ = \{( test visual samples of seen and unseen classes encountered so far ($X_{s_{te}}$ and $X_{u_{te}}$), their class labels ( $Y_{s_{te}}$ and $Y_{u_{te}}$), class attributes($A^t$))\}. Let $R^t$ be the replayed seen visual features of previous tasks. During the training phase of each task, data pertaining to current task and $R^t$ is used as training data. At the end of each task, the model's performance is evaluated by testing it on current as well as previous tasks' seen and unseen data. We operate in semantically transductive setting and do not use unseen visual features during training.
\vspace{-4pt}
\section {CGZSL: Proposed Methodology}
\vspace{-4pt}
\paragraph{Overall Framework.} The overall framework of our approach is shown in Fig. \ref{fig:my_label}. Given a time step $t$, we first train a generator $G$  which employs a cosine similarity based formulation enabling it to dynamically incorporate any number of categories or tasks over time (Sec. \ref{gan}). In order to ensure that generated visual features are discriminative w.r.t class distribution at current time step, we impose normalized discriminative loss (Sec. \ref{real_class}).
In addition, we propose to utilise incremental bi-directional alignment in order to adapt and ensure knowledge transfer from previous $(t-1)$ tasks and strengthen semantic relationship among classes encountered till time $t$ reducing catastrophic forgetting (Sec. \ref{iba}). At time step $(t+1)$ we use generative replay strategy  to generate seen class visual features ($R^t$) for all previous tasks and combine them with the seen class samples from the current task (Sec. \ref{gen_replay}). This procedure is repeated for the next time step.

\vspace{-3pt}
\subsection{Cosine Similarity-based GAN Classifier} 
\label{gan}
\vspace{-5pt}
 We learn a generative model which comprises of a generator, $G_{\theta}:\mathcal{Z}\times\mathcal{A}\to \mathcal{X}$ and a discriminator $D:\mathcal{A}\to \mathcal{X}$. The generator takes as input random noise, $z\in Z^{d}$ and class attributes $a$$\in$$A^t$ and outputs a generated visual feature belonging to the same class. On the other hand, the discriminator $D_{\phi}$ takes class attributes $a$$\in$$A^t$  as input and outputs identifier projection of the attribute. \textit{Identifier projection is a projection of the attribute in the visual space}. The $G_{\theta}$ and $D_{\phi}$ are trained adversarially where discriminator tries to minimize the cosine similarity between identifier projection and generated seen features belonging to same class, while the generator tries to maximize this cosine similarity. In addition to minimizing the aforementioned cosine similarity, the discriminator tries to maximize cosine similarity between real visual features and the corresponding identifier projection. Through this adversarial training, $G_{\theta}$ learns to generate visual features similar to the real visual features and $D_{\phi}$ learns a better mapping for the attributes. Training the $G_{\theta}$ and $D_{\phi}$ adversarially with the cosine similarities between generated seen features($X'$), identifier projection($D(a)$) and the real visual features ($X^{t}_{s_{tr}}$ $\cup$ $R^t$), is formulated as:
\vspace{-9pt}
\begin{equation}
\begin{split}
    L_{GAN} 
    &= \mathbb{E}_{x\sim p_{data}(X^{t}_{s_{tr}}\cup R^t)}\hspace{1mm}\left[\hspace{1mm} \log\hspace{1mm}\left[\hspace{1mm}\cos(x,D(a))\hspace{1mm}\right]\right]\\
    &+ \mathbb{E}_{x'\sim p_{\theta}(X'|(a))}\hspace{1mm}\left[\hspace{1mm}\log\hspace{1mm}\left[1 - \cos\hspace{1mm}(x',D\hspace{1mm})\hspace{1mm}\right]\right]
\label{eqn:1}
\end{split}  
\vspace{-3pt}
\end{equation}
where the distributions of real visual feature and generated visual feature are denoted by $p_{data}(X^{t}_{s_{tr}} \cup R^{t})$ and $p_{\theta} (X'|(a))$. 
During testing, we compute the cosine similarity between test sample and identifier projections of all the class attributes encountered so far. The test sample is assigned the class label of identifier projection with which it shares maximum similarity. Thus we are able to achieve continual generalized zero-shot classification using a single GAN model without the need for training a linear classifier during each task. The proposed architecture is simple and allows easy adaptation to increasing number of classes, as the classification is performed by merely computing cosine similarities between the identifier projections of the classes encountered so far and the test sample.

\vspace{-3pt}
\subsection{Real and Pseudo Normalized Loss}
\label{real_class}
\vspace{-5pt}
As new tasks or classes gradually arrive  over time, the specific features that allow us to best discriminate among the pool of classes dynamically changes. Thus, in order to ensure that the generated visual features are discriminative w.r.t the class distribution at any given time step $t$, we impose a set of losses on the generated visual features from current and all previous timesteps.
Specifically, we calculate softmax score of the cosine similarity between all identifier projections and visual features. The class label corresponding to identifier projection with highest softmax score is the predicted label. We use three classification losses- (i) Real classification loss ($L_{rcl}$) is the classification loss corresponding to real visual features (current task + replayed samples). 
$L_{rcl}$ enforces the discriminator to consider inter-class distances while mapping the attributes.
(ii) The classification loss corresponding to generated seen visual features called pseudo-visual classification loss ($L_{pcl}$) is added to the generator. $L_{pcl}$ encourages generator to generate more discriminative visual features. (iii) Since visual features of unseen classes are unavailable, a classification loss of generated unseen visual features called seen-normalized loss $L_{snl}$ is added to the discriminator. $L_{snl}$ serves as reference for finding appropriate mapping for unseen attributes. 

The classification losses $L_{rcl}$, $L_{pcl}$ and $L_{snl}$ are defined as follows:
\vspace{-8pt}
\begin{equation}
\begin{split}
L_{rcl}, L_{pcl}, L_{snl}
&= c\_e( \log \frac{\exp(\cos(x,D(a_{i})))}{\sum_{i\in A^{t}} \exp(\cos(x,D(a_{i})))}, y_{i} )
\label{eqn:2}
\vspace{-14pt}
\end{split}
\vspace{-14pt}
\end{equation}

\noindent $x$ corresponds to the real visual features in $L_{rcl}$, generated seen visual features in $L_{pcl}$ and generated unseen visual feature in $L_{snl}$. $c\_e$ stands for cross entropy and $y_{i}$ is the true class label of $x$. $t$ covers only seen classes in $L_{rcl}$ and $L_{pcl}$; but sums over seen and unseen classes for $L_{snl}$. During testing, classification spans all classes encountered till $t$.

\vspace{-3pt}
\subsection{Incremental Bi-directional Alignment Loss}
\label{iba}
\vspace{-5pt}
 Owing to the nature of the CGZSL setting, the visual feature space dynamically changes as new seen and unseen classes are added over time. Furthermore, the visual features for seen classes belonging to previous time steps and entire pool of unseen classes are not available during the current time step. Thus there is a need for a mechanism that can forward transfer knowledge from previous tasks to avoid catastrophic forgetting and exploit the current semantic structure to generate better visual features (especially unseen). To this end, we propose to use an incremental bi-directional alignment loss ($L_{iba}$) consisting of nuclear loss and semantic alignment loss. Semantic alignment loss helps in using semantic information \cite{lsrgan} as a reference for generating unseen visual features, while the nuclear loss aids in transferring visual information from seen classes to identifier projections (projection of attributes in visual space).

The visual similarity between two classes $c_{i}$ and $c_{j}$ is the cosine similarity between their class means. It is represented as $X_{sim}(\mu_{c_{i}} ;\mu_{c_{j}})$ where $\mu_{c_{i}}$ stands for the mean visual feature of class $i$. Let semantic similarity between the classes be represented as $\tau_{sim}(a_{c_{i}},a_{c_{j}})$ where $a_{c_{i}}$ is the attribute of class $i$. Semantic alignment loss constraints $X_{sim}(\mu_{c_{i}} ;\mu_{c_{j}})$ to lie in a range $\tau_{sim}(a_{c_{i}},a_{c_{j}})$ plus or minus $\epsilon$ (hyper-parameter), thus transferring the semantic structure to generated features. With the addition of new classes, the semantically similar classes of $c_i$ may change. Hence we incrementally calculate the semantic alignment loss ($L_{sal}$) for all classes encountered so far. Nuclear loss is the L2 norm between $\mu_{c_{i}}$ and corresponding identifier projection.
 
The incremental bi-directional semantic alignment loss is given by:
\vspace{-11pt}
\begin{equation}
\begin{split}
    L_{sal}
    &= \min_{\theta_{g}} \frac{1}{N}\sum^{N}_{i=1}\sum_{j\in I_{c_{i}}} ||\max (0,X_{sim}(\mu_{c_{j}}, \mu'_{c_{i}})\\
    &- (\tau_{sim}(a_{c_{j}},a_{c_{i}}) + \epsilon))||^{2} + \\
    &||\max (0,(\tau_{sim}(a_{c_{j}},a_{c_{i}})- \epsilon) -X_{sim}(\mu_{c_{j}}, \mu'_{c_{i}}))||^{2}
    \label{eqn:6}
\end{split}
\vspace{-17pt}
\end{equation}
\begin{equation}
\begin{split}
L_{nuclear}=||\mu_{c_{i}}-S_{c_{i}}||^{2}
    \label{eqn:7}
\end{split}
\vspace{-12pt}
\end{equation}

\begin{algorithm}[H]
\footnotesize
    \caption{Proposed CGZSL Method}\label{your_label}
    \begin{algorithmic}
        \STATE \textbf{Input\hspace{5mm}:}\hspace{5mm}$D^t_{tr}$, $D^t_{te}$, $G$, $D$\\
        \textbf{Output\hspace{5mm}:}\hspace{5mm}Predicted labels $y_{pred}$\\
        \textbf{Parameters\hspace{5mm}:}\hspace{5mm}$\theta_{G}$, $\phi_{D}$\\
        \IF{new task arrived}
            \IF{task number greater than 1}
                  \STATE $R^{t}$ =  replay data till task t-1; $X^{t}_{s_{tr}}$ = $X^{t}_{s_{tr}}$\hspace{1mm}$\cup$\hspace{1mm}$R^{t}$
            \ENDIF
            \FOR{epochs = $1$ to $N$}
                    \STATE {$X_{s}^{'}$ = $G(z,A_{s}^t)$}  \hspace{1mm} \texttt{// seen pseudo-visual features}
                    \STATE $X_{u}^{'}$ = $G(z,A_{u}^t)$ \hspace{0.1mm} \tt{// unseen pseudo-visual features}
                    \STATE  $L_{GAN}$ $\leftarrow$ eqn (\ref{eqn:1}) \texttt{// using $\tt{X}^{t}_{s_{tr}}$, $\tt{X}_{s}^{'}$ and $\tt{D}(A_{s}^{t})$ }
                    \STATE  $L_{rcl}$ $\leftarrow$ eqn (\ref{eqn:2}) \hspace{1mm} \texttt{// using  $\tt{X}^{t}_{s_{tr}}$ and $\tt{D}(A_s^t)$}
                    \STATE $L_{snl}$ $\leftarrow$ eqn ({\ref{eqn:2}}) \hspace{2mm} \texttt{// using $\tt{X}_{u}^{'}$, $\tt{D}(A^t)$ }
                    \STATE {// Overall $\tt{D}$ loss}
                    \STATE $L_{D}^{t}$ = $\lambda_{1}$ $L_{GAN}$ + $\lambda_{2}$ $L_{rcl}$ + $\lambda_{3}$ $L_{snl}$ 
                    \STATE $\phi_{D}$ = $\phi_{D}$ - $\eta_{1}$ $\times$ $\nabla$ $L_{D}^{t}$ \hspace{1mm} \texttt{// update $\tt{D}$}
                    \STATE $L_{pcl}$ $\leftarrow$ eqn ({\ref{eqn:2}}) \hspace{2mm} \texttt{// using $\tt{X}_{s}^{'}$ and $\tt{D}(A_s^t)$}
                    \STATE $L_{iba}$ $\leftarrow$ eqn ({\ref{eqn:6}}) and eqn (\ref{eqn:7}) \hspace{2mm} \texttt{// using $\tt{X}_{s}^{'}$  and $\tt{X}_{u}^{'}$}
                    \STATE \texttt{// Overall $\tt{G}$ loss}
                    \STATE $L_{G}^{t}$ = $\lambda_{1}$ $L_{GAN}$ + $\lambda_{2}$ $L_{pcl}$ +  $\lambda_{4}$ $L_{iba}$  
                    \STATE $\theta_{G}$ = $\theta_{G}$ - $\eta_{2}$ $\times$ $\nabla$ $L_{G_{s}}^{t}$ \hspace{1mm} \texttt{// Update $\tt{G}$}
            \ENDFOR
            \newline
          Inference: model is evaluated on $D_{te}^{t}$
        \ENDIF
    \end{algorithmic}
\end{algorithm}
\noindent where $N$ is the number of classes encountered so far. For a given class $c_{i}$, $S_{c_{i}}$ is the identifier projection, $\mu_{c_{i}}$ is the mean of real visual features and $\mu'_{c_{i}}$ is mean of generated visual features. Mean of real visual features is available only for seen classes. Let $I_{c_{i}}$ represent the set of $n_{c}$ nearest neighbours of $c_{i}$.

\vspace{-3pt}
\subsection{Generative Replay}
\label{gen_replay}
\vspace{-5pt}
We work in a setting where data arrives incrementally, and samples from previous tasks are not accessible during the current task.  This results in catastrophic forgetting.  In order to retain previously learned knowledge and adopt new knowledge, we use generative replay. Visual features of previously seen classes are generated by passing a concatenation of attribute and noise vectors to the generator network. A combination of generated features of previous tasks and real features from the current task act as input data to train the model.  To ensure that we are replaying credible and good quality visual features at a time step $t$,  we classify the generated visual features and replay only features that are classified correctly.
\vspace{-12pt}
\begin{equation}
\begin{split}
   R^{t} = G (z,A^{\leq t-1}_{s})
    \label{eqn7}
\end{split}
\vspace{-12pt}
\end{equation}
where $z$ $\sim$ $\mathcal N(0,1)$ and $A^{\leq t-1}_s$ denotes attributes of all the seen classes encountered so far.
\vspace{-3pt}
\subsection{Training and Inference}
\vspace{-4pt}
For a given time step $t(>1)$, we concatenate the replayed visual features and current task’s visual features from seen classes along with their corresponding labels. The concatenated data acts as input for training the generative model. During training, the discriminator and generator are trained sequentially on different tasks. After the training process, the discriminator learns a mapping function to map attributes to visual space. The generator learns to generate synthetic visual features conditioned on attributes. 
During testing, we classify the test sample using cosine similarity as described in Sec. \ref{real_class}. Based on the setting we are working in, accuracies are computed as described in Appendix.

\definecolor{Gray}{gray}{0.9}
\newcolumntype{g}{>{\columncolor{Gray}}c}
\begin{table*}
\centering
\resizebox{16cm}{!}{%
   \begin{tabular}{{lllgllgllgllgllgllllll}}
    \toprule
      &
      \multicolumn{3}{c}{aPY} &
      \multicolumn{3}{c}{AWA1} &
      \multicolumn{3}{c}{AWA2} &
      \multicolumn{3}{c}{CUB} &
      \multicolumn{3}{c}{SUN} \\
      & {mSA} & {mUA} & {mH} & {mSA} & {mUA} & {mH}& {mSA} & {mUA} & {mH}& {mSA} & {mUA} & {mH}& {mSA} & {mUA} & {mH} \\
      \midrule
    \multicolumn{16}{c}{\textbf{Online-CGZSL}}\\
    \hline
    EWC \cite{ewc} &40.44&-&-&53.62&-&-&55.18&-&-&38.64&-&-&25.79&-&-\\
    AGEM \cite{agem} &-&-&-&-&-&-&57.25&-&-&41.93&-&-&28.61&-&-\\
    Seq-fCLSWGAN \cite{fclswgan}  &26.92&16.34&20.34&24.83&15.29&18.93&24.56&16.84&19.98&14.26&10.0&11.76&14&10.38&11.92\\
    Seq-CVAE \cite{cvae}  &63.58& 16.24&25.87&72.41&26.39&38.68&71.84&25.33&37.45&36.23&21.70& 27.14&30.87&22.69&26.16\\
    Seq-CADA \cite{cada} &62.54&22.89&33.51&75.45&23.28&45.58&78.59&35.81&49.20&53.18&27.22&36.00&40.67&24.13&30.29\\
    CV +mem \cite{cvae}  &77.74&27.62&40.76&\textbf{81.35}& 39.63&53.29&\textbf{85.29}&34.64&49.27&64.27&26.39&37.42&36.11&22.39&27.64\\
    CA +mem \cite{cada} &67.18&35.06& 46.07&79.23&58.28&67.15&78.49&55.32&64.90&64.83&40.99&50.22&45.25&28.19&34.74\\
    A-CZSL \cite{azsl} &59.44&18.94&28.51&\textbf{78.56}&37.41&52.23&79.75&38.11&51.57&41.77&24.34&30.39&16.33&9.46&11.98\\
    Tf-GCZSL \cite{iisc2}  
    &69.01&24.81&37.59&65.01&50.43&56.80&65.59&60.17&62.76&\textbf{65.36}&44.19&52.80&40.18&41.05&37.81\\
    DVGR-CZSL \cite{ghosh} &57.97&28.46&35.27&73.55&44.21&55.22&78.36&40.75&53.62&45.68&16.65&23.94&23.02&14.00&16.92\\
    NM-ZSL \cite{cln} &-&-&-&-&-&-&84.25&48.94&61.91&61.40&49.64&\textbf{54.9}&\textbf{53.57}&27.51&36.35\\
     \hline
    Ours &\textbf{75.79}&\textbf{37.72}&\textbf{49.11}&78.99& \textbf{69.33}&\textbf{73.33}&80.06&\textbf{72.62}&\textbf{75.44}&44.09&\textbf{53.57}&48.03&41.87&\textbf{45.72}&\textbf{43.20}\\
    \bottomrule
  \end{tabular}
  }
\vspace{-4pt}
   \caption{\footnotesize Mean seen accuracy (mSA), mean unseen accuracy (mUA), and their harmonic mean (mH) for Online-CGZSL setting.}
   \vspace{-10pt}
\label{table1}
\end{table*}

\begin{table}
\centering
\footnotesize
  \begin{tabular}{{lllll}}
    \toprule
      &
      \multicolumn{2}{c}{Dynamic-CGZSL} &
      \multicolumn{2}{c}{Online-CGZSL} \\
      \hline
      & {AWA2} & {CUB} & {AWA2} & {CUB} \\
      \midrule
    A-CZSL &0.09&0.14&0.07&0.13\\
    Tf-GCZSL    
    &0.09&0.07&0.09&0.09\\
    DVGR-CZSL &0.15&0.14&0.12&0.12\\
    NM-ZSL &0.11&0.08&0.10&0.08\\
     \hline
    Ours &0.09&0.13&0.09&0.11\\
    \bottomrule
  \end{tabular}
\vspace{-4pt}
  \caption{\footnotesize Forgetting measure for dynamic \& online settings on AWA2 \& CUB datasets.}
  \vspace{-10pt}
\label{forgettingtable}
\end{table}

\begin{table*}
\centering
\resizebox{16cm}{!}{%
   \begin{tabular}{{lllgllgllgllgllgllllll}}
    \toprule
      &
      \multicolumn{3}{c}{aPY} &
      \multicolumn{3}{c}{AWA1} &
      \multicolumn{3}{c}{AWA2} &
      \multicolumn{3}{c}{CUB} &
      \multicolumn{3}{c}{SUN} \\
      & {mSA} & {mUA} & {mH} & {mSA} & {mUA} & {mH}& {mSA} & {mUA} & {mH}& {mSA} & {mUA} & {mH}& {mSA} & {mUA} & {mH} \\
      \midrule
    \multicolumn{16}{c}{\textbf{Dynamic-CGZSL}}\\
    \hline
    EWC\cite{ewc}&42.68&-&-&50.41&-&-&53.29&-&-&40.95&-&-&24.46&-&-\\
    AGEM\cite{agem} &-&-&-&-&-&-&57.0&-&-&43.1&-&-&27.15&-&-\\
    Seq-fCLSWGAN\cite{fclswgan} 
    &25.73&15.27&19.17&23.39& 12.57&16.35&22.88&14.92&18.06&13.14&7.47&9.53&12.70&8.63&10.27\\
    Seq-CVAE\cite{cvae}   &65.87& 17.90&  25.84&70.24&28.36& 39.32&73.71&26.22&36.30&38.95&20.89&27.19&29.06&21.33&24.33\\
    Seq-CADA\cite{cada}  &61.17& 21.13&  26.37&78.12&25.93 &47.06&79.89&36.64&47.99&55.55&26.96&35.62&42.21&23.47&29.60\\
    CV +mem\cite{cvae}    &78.15&28.10& 40.21&\textbf{85.01}& 37.49& 51.60&88.36&33.24&47.89&63.16&27.50&37.84&37.50&24.01&29.15\\
    CA +mem\cite{cada}    &66.30&\textbf{36.59}& 45.08&81.86&61.39&69.92&82.19&55.98&65.95&\textbf{68.18}&42.44&50.68&47.18&30.30&34.88\\
    A-CZSL\cite{azsl} &64.06&16.82&24.46&78.03&35.38&52.51&82.91&42.19&57.74&47.34&27.67&32.77&15.26&8.93&11.27\\
    Tf-GCZSL\cite{iisc2}    
    &72.12& 19.66& 30.90&61.79& 57.77& 59.72&67.42& 58.08&62.41&44.52&43.21&43.85&27.76& 39.09&32.46\\
    DVGR-CZSL\cite{ghosh} &69.67&31.58&43.44&76.9&42.04&53.38&78.17&40.44&50.89&47.28&22.58&29.77&23.37&16.33&18.65\\
    NM-ZSL\cite{cln} &\textbf{79.60}&22.29&  32.61&75.59&60.87& 67.44&\textbf{89.22}&51.38& 63.41&64.91&46.05&\textbf{53.79}&\textbf{50.56}& 35.55&41.65\\
     \hline
    Ours  &74.92&33.94&\textbf{46.26}& 79.51& \textbf{69.13}& \textbf{73.49}&79.19& \textbf{70.71}& \textbf{74.09}&50.51&\textbf{52.2}& 51.18 &45.43&\textbf{44.59}&\textbf{44.56}\\
    \hline  
      
    \multicolumn{16}{c}{\textbf{Static-CGZSL}}\\
    \hline
    EWC \cite{ewc} &37.65&-&-&42.6&-&-&43.71&-&-&35.89&-&-&21.36&-&-\\
    AGEM \cite{agem} &-&-&-&-&-&-&52.61&-&-&39.54&-&-&24.22&-&-\\
    Seq-fCLSWGAN \cite{fclswgan} &21.18&8.96&12.59&20.32&9.25&12.71&18.97&8.45&11.69&11.16&6.54&8.25&10.02& 6.89&8.16\\
    Seq-CVAE \cite{cvae}  &51.57&  11.38&  18.33&59.27 & 18.24& 27.14&61.42&  19.34 &28.67& 24.66& 8.57&  12.18&16.88& 11.40& 13.38\\
    Seq-CADA \cite{cada}  &45.25&  10.59&  16.42&51.57&  18.02& 27.59&52.30&  20.30& 30.38& 40.82& 14.37& 21.14&25.94& 16.22& 20.10\\
    CV +mem\cite{cvae}   &\textbf{64.88}& 15.24&  23.90&78.56& 23.65 &35.51&\textbf{80.97}& 25.75& 38.34& 44.89&13.45& 20.15&23.99& 14.10& 17.63\\
    CA +mem\cite{cada}   &57.69& 20.83&  28.84&62.64& 38.41&45.38&62.80&39.23&46.22& 43.96&32.77& 36.06&27.11&21.72&22.92\\
    A-CZSL \cite{azsl} &58.14&15.91&23.05&71.00&24.26&35.75&70.16&25.93&37.19&34.47&12.00&17.15&-&-&-\\
    Tf-GCZSL \cite{iisc2}  &57.92&21.22&29.55&64.00&38.34&46.16&64.89&40.23&48.33&46.63&32.42&36.31&28.09&24.70&24.79\\
    DVGR-CZSL \cite{ghosh} &62.5&22.7&31.7&65.1&28.5&38&73.5&28.8&40.6&44.87&14.55&21.66&22.36&10.67&14.54\\
    NM-ZSL \cite{cln}  &45.26 &21.35& 27.18&70.90& 37.46& 48.75&76.33& 39.79& 51.51& \textbf{55.45}&\textbf{43.25}& \textbf{47.04}&\textbf{50.01}&19.77 &\textbf{28.04}\\
    \hline
    Ours  &60.96& \textbf{26.5}& \textbf{33.96} &66.30&\textbf{50.63}& \textbf{55.59}  &67.15& \textbf{54.19}& \textbf{61.32}&42.39& 37.09& 36.35  &21.44 &\textbf{27.57} &23.87\\

    \bottomrule
  \end{tabular}
  }
  \vspace{-4pt}
   \caption{\footnotesize Mean seen accuracy (mSA), mean unseen accuracy (mUA), and their harmonic mean (mH) for dynamic \& static-CGZSL settings.}
   \vspace{-10pt}
\label{table2}
\end{table*}

\vspace{-4pt}
\section{Experiments and Results}
\label{sec_experiments}
\vspace{-4pt}
For each of the three CGZSL problem settings -- static, dynamic and online, we evaluate the proposed approach on five different benchmark datasets: Animals with Attributes (AWA1 and AWA2) \cite{awa,zslcompre}, Attribute Pascal and Yahoo (aPY)\cite{apy},  Caltech-UCSD-Birds 200-2011 (CUB) \cite{cub} and SUN \cite{sun}. We follow the data splits mentioned in \cite{ghosh,iisc} for fair comparison. We follow \cite{cln,iisc,iisc2} and evaluate our model on mean seen accuracy (mSA), mean unseen accuracy (mUA) and mean harmonic value(mH) over all tasks. Further details on the metrics are provided in the Appendix.

\begin{figure}[h]
    \centering
    \includegraphics[width=0.3\textwidth]{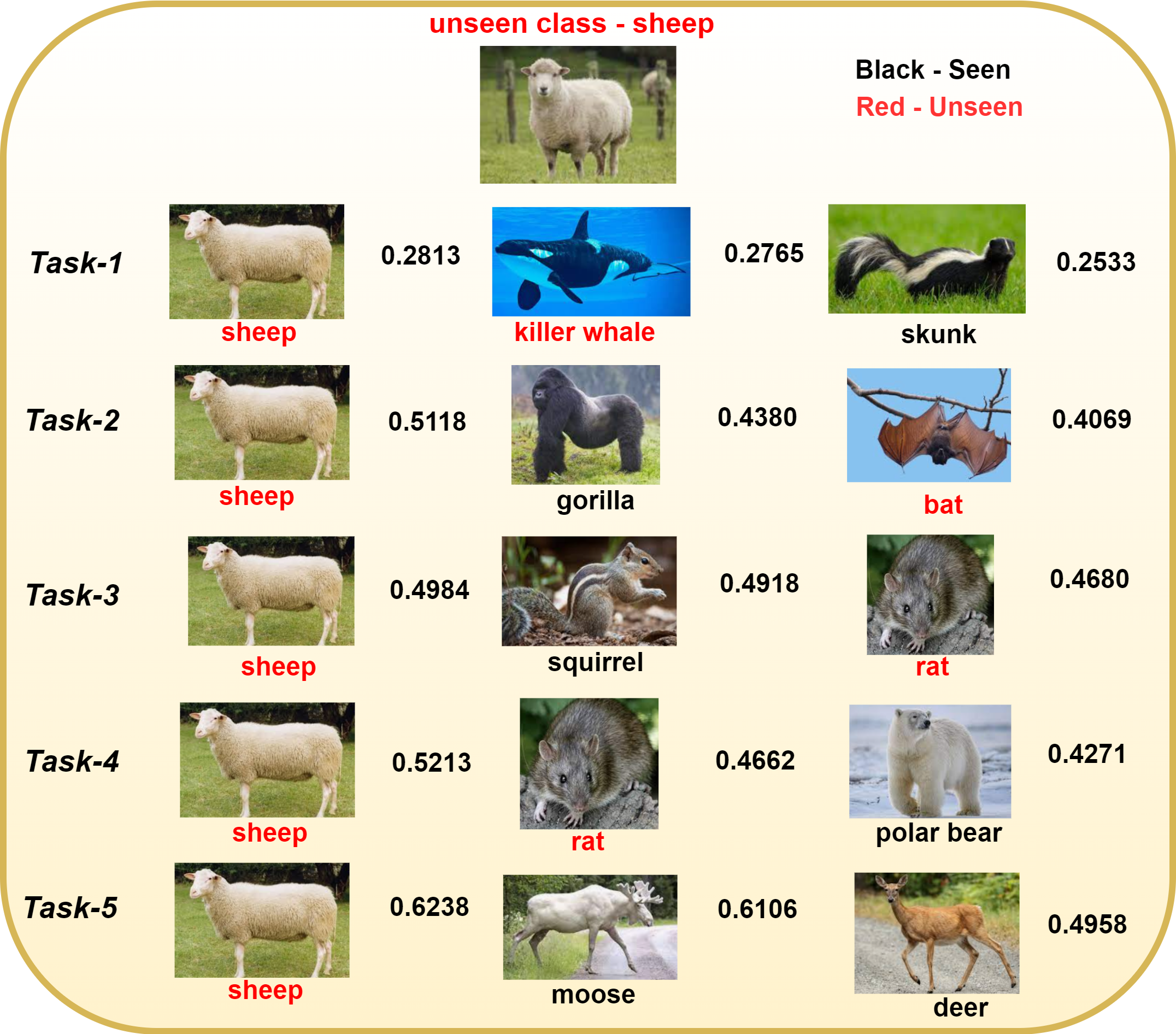}
    \vspace{-2pt}
    \caption{\footnotesize Cosine similarity scores of unseen class `sheep’ of AWA2 dataset w.r.t. identifier projection. Top three cosine similarity scores shared with `sheep’ during inference at every task is shown. Unseen classes are in red, seen classes are in black. For detailed explanation, refer appendix.}
    \vspace{-12pt}
    \label{qualiresult}
\end{figure}
\begin{figure}
    \centering
    \includegraphics[width=0.8\linewidth]{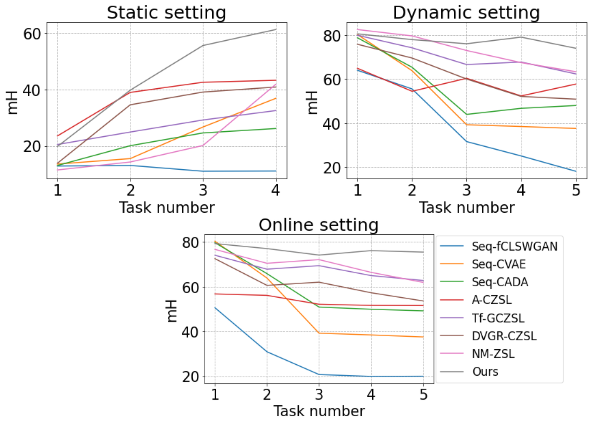}
    \vspace{-8pt}
    \caption{\footnotesize Task-wise mH values for AWA2 dataset in static (top left), dynamic (top right) and online (below) setting.}
    \vspace{-10pt}
    \label{taskwise}
\end{figure}

\vspace{-11pt}
\paragraph{Baselines.}
In order to analyze the need for a CGZSL model for tackling dynamic addition of new classes, we compare our work with sequential versions of well-known traditional GZSL baselines: f-CLSWGAN \cite{fclswgan}, CVAE \cite{cvae} and CADA \cite{cada} (which we call Seq-fCLSWGAN, Seq-CVAE and Seq-CADA respectively). Existing continual learning methods have shown acceptable performance in retaining previously learned knowledge, but fail to handle unseen classes. We compare our method with notable continual learning frameworks EWC \cite{ewc} and A-GEM \cite{agem}. Considering the importance of memory based replay for overcoming catastrophic forgetting, we assess our model with sequential versions of CVAE and CADA after adding a memory based replay strategy (which we call CV+mem and CA+mem respectively). We further evaluate the performance of our model against recent efforts that try to solve the CGZSL problem: (i) A-CZSL \cite{azsl}; (ii) Tf-GCZSL \cite{iisc2}; (iii) DVGR-CZSL \cite{ghosh}; (iv) NM-ZSL \cite{cln}. We compare these baselines with our model on the three settings described in Sec. \ref{cgzslwhat}.

\vspace{-10pt}
\paragraph{Results.}
Table \ref{table1} shows the performance comparison of our model in online setting with multiple baselines. Table \ref{table2} shows the performance of our model on dynamic and static settings. 
As a first effort in addressing the \underline{Online-CGZSL} setting, we convert an unseen class from the $t-1^{th}$ task to a seen class in the $t^{th}$ task, in addition to introducing new seen and unseen classes, and use this for evaluating our method as well as the baselines. We introduce benchmark results on all datasets and baselines for this setting in Table \ref{table1}. We observe that our model outperforms the baselines by a significant margin of 3.04\%, 6.18\%, 10.54\% and 5.39\% on aPY, AWA1, AWA2 and SUN datasets respectively on mean harmonic value(mH). In Fig. \ref{taskwise} (bottom), we show mH per task for the AWA2 dataset in online setting.

\noindent On \underline{Dynamic-CGZSL} (Table \ref{table2}), we observe that our model yielded 1.18\%, 3.57\%, 8.14\% and 2.91\% increase in mH on aPY, AWA1, AWA2 and SUN datasets and competitive performance on CUB dataset with respect to all baselines. Our model outperforms 4 out of 5 state-of-the-art unseen accuracies on CUB, AWA1, AWA2 and SUN datasets. Considering unseen class accuracy poses a greater challenge than seen classes, the consistent performance of the proposed approach on unseen accuracy corroborates our claim. We believe that adding other continual learning strategies (such as regularizers) can further improve the performance of our method on seen classes. In
Fig. \ref{taskwise} (top right), we evaluate mH per task for the AWA2 dataset in dynamic setting.

\noindent On \underline{Static-CGZSL} (Table \ref{table2}), we observe that our model obtained state-of-the art performance with noticeable gains of  2.26\%, 6.84\% and 9.81\% over the baselines on aPY, AWA1 and AWA2 datasets for mH. The results support our claim that our model performs exceedingly well on unseen accuracies across all settings, thanks to our bi-directional alignment loss which greatly helped in boosting performance. Fig. \ref{taskwise} (top left) shows mH of the model versus the task number for the AWA2 dataset in static setting.

We report forgetting measures of AWA2 and CUB datasets in dynamic and online settings in Table \ref{forgettingtable}, which shows that addition of generative replay helps to reduce forgetting. Fig. \ref{qualiresult} demonstrates how cosine similarity of an unseen class changes across different tasks during inference.

\begin{table}
\resizebox{8cm}{!}{%
   \begin{tabular}{{lllllllllll}}
    \toprule
      &
      \multicolumn{3}{c}{AWA1} &
      \multicolumn{3}{c}{CUB}\\
      & {mSA} & {mUA} & {mH} & {mSA} & {mUA} & {mH} \\
      \midrule
    Ours &78.99&69.33&73.33&44.09&53.57&48.03\\
    Ours w/o replay &21.27&33.64&20.24&1.89&2.74&2.21\\
    Ours w/o $L_{sal}$ &78.16&64.38&69.10&33.83&49.62&40.10\\
    Ours w/o $L_{nuclear}$ &72.33&67.62&69.31&37.66&42.36&39.53\\
    Ours w/o $L_{sal}$ and $L_{nuclear}$ &57.39&60.22&56.38 &14.14&36.52&19.88\\
    Ours w/o  $L_{rcl}$ and $L_{pcl}$ &78.65&33.4&45.65&13.79&9.50&10.72\\
    Ours w/o  $L_{snl}$ &77.16&61.87&67.74&37.53&45.66&40.92\\
    Ours w/o  replay, $L_{rcl}$ and $L_{pcl}$ &3.36&6.29&4.29&0.53&0.81&0.58\\
    \bottomrule
  \end{tabular}
  }
  \vspace{-4pt}
  \caption{\footnotesize Ablation study on AWA1 \& CUB datasets for online-CGZSL setting.}
 
  \vspace{-15pt}
\label{table:4}
\end{table}

\vspace{-4pt}
\section{Ablation Studies and Analysis}
\vspace{-8pt}
To understand the importance of different components, Table \ref{table:4} shows ablation studies of our approach on coarse and fine-grained datasets i.e AWA1 and CUB.

\noindent \textbf{Role of replay data.} In order to retain previous task information, we replay samples from seen classes using the generator network. To confirm that replaying data helps in overcoming catastrophic forgetting we execute the model without replaying visual features of seen class and observe that mean harmonic accuracy value drops by 53.09\% on AWA1 and 45.82\% on CUB datasets respectively.

\noindent \textbf{Effect of bi-directional alignment.}
We hypothesized that the bi-directional alignment loss helps the model accommodate and align new classes; we empirically show its significance by validating our model with and without the loss. Having only one of the alignment (one-directional loss) i.e either $L_{sal}$ or $L_{nuclear}$ resulted in the drop of performance by 4\% on AWA1 and 8\% on CUB datasets. The performance of the model without bi-directional loss significantly drops by 16.95\% and 28.15\% on AWA1 and CUB datasets which supports the need for it.

\noindent \textbf{Effect of classification loss.}
The absence of the classification losses which catalyzes the bi-directional alignment clearly indicates a drop of 27.68\% on AWA1 and 37.31\% on CUB datasets.

\vspace{-8pt}
\section{Conclusions and Future Work}
\vspace{-6pt}
Existing literature has witnessed accelerated efforts towards solving the GZSL problem, but the static nature of the setting does not allow gradual addition of categories (seen or unseen) over time which hampers applicability in the real-world. In this work, we address the aforementioned issue, and push the existing continuous generalized zero-shot learning (CGZSL) boundaries towards a more pragmatic \emph{Online-CGZSL} setting. To this end, we propose a bi-directional alignment-based generative framework that can tackle changes in pool of categories and data availability, over time. Extensive experiments conducted on benchmark datasets verify that our approach demonstrates superior performance compared to existing baselines. Our future efforts will include exploring dynamic feature-based attention mechanisms for designing adaptable models to better address the CGZSL setting.

\noindent \textbf{Limitation and Broader Impacts.}
The continual generalized zero-shot setting (CGZSL) is a step towards designing AI systems that are adaptable to changes in data availability and dynamic additions of object categories, which inevitably occur in the real-world over time. However, getting annotated attributes for all datasets may be non-trivial. Our work has no known social detrimental impacts other than the ones generally associated with the development of a new AI system.

\noindent \textbf{Acknowledgements.} This work has been partly supported by the funding received from DST through the IMPRINT and ICPS programs. Hari Chandana Kuchibhotla would like to thank MoE for the PMRF fellowship support. We thank the anonymous reviewers for their valuable feedback that improved the presentation of this paper.

{\small
\bibliographystyle{ieee_fullname}
\bibliography{cgzsl}

\begin{thebibliography}{10}\itemsep=-1pt

\bibitem{shivam}
Shivam Chandhok and Vineeth~N Balasubramanian.
\newblock {Two-level adversarial visual-semantic coupling for generalized
  zero-shot learning.}
\newblock In {\em IEEE Winter Conference on Applications of Computer Vision
  (WACV)}, pages 3100--3108, 2021.

\bibitem{agem}
Arslan Chaudhry, Marc’Aurelio Ranzato, Marcus Rohrbach, and Mohamed
  Elhoseiny.
\newblock Efficient lifelong learning with a-{GEM}.
\newblock In {\em International Conference on Learning Representations (ICLR)},
  2019.

\bibitem{apy}
Ali Farhadi, Ian Endres, Derek Hoiem, and David Forsyth.
\newblock Describing objects by their attributes.
\newblock In {\em IEEE Conference on Computer Vision and Pattern Recognition
  (CVPR)}, pages 1778--1785. IEEE, 2009.

\bibitem{devise}
Andrea Frome, Greg~S. Corrado, Jonathon Shlens, Samy Bengio, Jeffrey Dean,
  Marc’Aurelio Ranzato, and Tomas Mikolov.
\newblock {Devise: A deep visual-semantic embedding model.}
\newblock In {\em Advances in Neural Information Processing Systems (NIPS)},
  page 2121–2129, 2013.

\bibitem{zerovaegan}
Rui Gao, Xingsong Hou, Jie Qin, Li Liu, Fan Zhu, and Ling Shao.
\newblock {“Zero-VAE-GAN: Generating unseen features for generalized and
  transductive zero-shot learning.}
\newblock In {\em IEEE Transactions on Image Processing (TIP)}, pages
  3665--3680, 2020.

\bibitem{iisc}
Chandan Gautam, Sethupathy Parameswaran, Ashish Mishra, and Suresh Sundaram.
\newblock Generalized continual zero-shot learning, 2021.

\bibitem{iisc2}
Chandan Gautam, Sethupathy Parameswaran, Ashish Mishra, and Suresh Sundaram.
\newblock Tf-gczsl: Task-free generalized continual zero-shot learning.
\newblock In {\em arXiv preprint arXiv:2103.10741}, 2021.

\bibitem{azsl}
Subhankar Ghosh.
\newblock Adversarial training of variational auto-encoders for continual
  zero-shot learning(a-czsl).
\newblock In {\em International Joint Conference on Neural Networks (IJCNN)},
  pages 1--8, 2021.

\bibitem{ghosh}
Subhankar Ghosh.
\newblock Dynamic vaes with generative replay for continual zero-shot learning,
  2021.

\bibitem{ewc}
James Kirkpatrick, Razvan Pascanu, Neil Rabinowitz, Joel Veness, Guillaume
  Desjardins, ..., and Raia Hadsell.
\newblock {Overcoming catastrophic forgetting in neural networks.}
\newblock In {\em Proceedings of the national academy of sciences}, pages
  3521--3526, 2017.

\bibitem{awa}
Christoph~H Lampert, Hannes Nickisch, and Stefan Harmeling.
\newblock Learning to detect unseen object classes by between-class attribute
  transfer.
\newblock In {\em IEEE Conference on Computer Vision and Pattern Recognition
  (CVPR)}, pages 951--958. IEEE, 2009.

\bibitem{zeroshotvis}
Christoph~H. Lampert, Hannes Nickisch, and Stefan Harmeling.
\newblock {Attribute based classification for zero-shot visual object
  categorization.}
\newblock In {\em IEEE Transactions on Pattern Analysis and Machine
  Intelligence (TPAMI)}, pages 453--465, 2013.

\bibitem{lisgan}
Jingjing Li, Mengmeng Jing, Ke Lu, Zhengming Ding, Lei Zhu, and Zi Huang.
\newblock {Leveraging the invariant side of generative zero-shot learning..}
\newblock In {\em IEEE Conference on Computer Vision and Pattern Recognition
  (CVPR)}, pages 7402--7411, 2019.

\bibitem{lwf}
Zhizhong Li and Derek Hoiem.
\newblock {Learning without forgetting.}
\newblock In {\em IEEE Transactions on Pattern Analysis and Machine
  Intelligence (TPAMI)}, pages 2935--2947, 2017.

\bibitem{genclassinc}
Xialei Liu, Chenshen Wu, Mikel Menta, Luis Herranz, Bogdan Raducanu, ..., and
  Joost van~de Weijer.
\newblock {Generative feature replay for class-incremental learning.}
\newblock In {\em IEEE Conference on Computer Vision and Pattern Recognition
  (CVPR) workshops}, pages 226--227, 2020.

\bibitem{openlongtailrecognition}
Ziwei Liu, Zhongqi Miao, Xiaohang Zhan, Jiayun Wang, Boqing Gong, and Stella~X.
  Yu.
\newblock Large-scale long-tailed recognition in an open world.
\newblock In {\em IEEE Conference on Computer Vision and Pattern Recognition
  (CVPR)}, 2019.

\bibitem{packnet}
Arun Mallya and Svetlana Lazebnik.
\newblock {PackNet: Adding Multiple Tasks to a Single Network by Iterative
  Pruning.}
\newblock In {\em IEEE Conference on Computer Vision and Pattern Recognition
  (CVPR)}, pages 7765--7773, 2018.

\bibitem{clbayes}
Nikhil Mehta, Kevin Liang, Vinay~Kumar Verma, and Lawrence Carin.
\newblock {Continual Learning using a Bayesian Nonparametric Dictionary of
  Weight Factors.}
\newblock In {\em International Conference on Artificial Intelligence and
  Statistics (AISTATS)}, pages 100--108, 2021.

\bibitem{catasinfer}
MichaelMcCloskey and Neal J.Cohen.
\newblock {Catastrophic interference in connectionist networks.}
\newblock In {\em Psychology of learning and motivation, vol. 24 academic
  press}, pages 109--165, 1989.

\bibitem{cvae}
Ashish Mishra, Shiva~Krishna Reddy, Anurag Mittal, and Hema~A. Murthy.
\newblock {A generative model for zero shot learning using conditional
  variational autoencoders.}
\newblock In {\em IEEE Conference on Computer Vision and Pattern Recognition
  (CVPR) workshops}, pages 2188--2196, 2018.

\bibitem{sun}
Genevieve Patterson and James Hays.
\newblock Sun attribute database: Discovering, annotating, and recognizing
  scene attributes.
\newblock In {\em IEEE Conference on Computer Vision and Pattern Recognition
  (CVPR)}, pages 2751--2758. IEEE, 2012.

\bibitem{icarl}
Sylvestre-Alvise Rebuffi, Alexander Kolesnikov, G. Sperl, and Christoph~H.
  Lampert.
\newblock icarl: Incremental classifier and representation learning.
\newblock {\em 2017 IEEE Conference on Computer Vision and Pattern Recognition
  (CVPR)}, pages 5533--5542, 2017.

\bibitem{simplezsl}
Bernardino Romera-Paredes and Philip Torr.
\newblock {An embarrassingly simple approach to zero-shot learning.}
\newblock In {\em International Conference on Machine Learning (ICML)}, pages
  2152--2161, 2015.

\bibitem{pnn}
Andrei~A Rusu, Neil~C Rabinowitz, Guillaume Desjardins, Hubert Soyer, James
  Kirkpatrick, Koray Kavukcuoglu, Razvan Pascanu, and Raia Hadsell.
\newblock Progressive neural networks.
\newblock {\em arXiv preprint arXiv:1606.04671}, 2016.

\bibitem{cada}
Edgar Schonfeld, Sayna Ebrahimi, Samarth Sinha, Trevor Darrell, and Zeynep
  Akata.
\newblock {Generalized zero-and few-shot learning via aligned variational
  autoencoders.}
\newblock In {\em IEEE Conference on Computer Vision and Pattern Recognition
  (CVPR)}, pages 8247--8255, 2019.

\bibitem{cln}
Ivan Skorokhodov and Mohamed Elhoseiny.
\newblock Class normalization for (continual)? generalized zero-shot learning.
\newblock In {\em International Conference on Learning Representations (ICLR)},
  2021.

\bibitem{lsrgan}
Maunil~R. Vyas, Hemanth Venkateswara, and Sethuraman Panchanathan.
\newblock {Leveraging Seen and Unseen Semantic Relationships for Generative
  Zero-Shot Learning.}
\newblock In {\em European Conference on Computer Vision (ECCV)}, pages 70--86,
  2020.

\bibitem{cub}
C. Wah, Steve Branson, P. Welinder, P. Perona, and Serge~J. Belongie.
\newblock The caltech-ucsd birds-200-2011 dataset.
\newblock In {\em .} California Institute of Technology, 2011.

\bibitem{bookworm}
Kai Wang, Luis Herranz, Anjan Dutta, and Joost van~de Weijer.
\newblock Bookworm continual learning: beyond zero-shot learning and continual
  learning, 2020.

\bibitem{lzsl}
Kun Wei, Cheng Deng, and Xu Yang.
\newblock { Lifelong zero-shot learning.}
\newblock In {\em International Joint Conference on Artificial Intelligence
  (IJCAI)}, pages 551--557, 2020.

\bibitem{zslcompre}
Yongqin Xian, Christoph~H Lampert, Bernt Schiele, and Zeynep Akata.
\newblock Zero-shot learning—a comprehensive evaluation of the good, the bad
  and the ugly.
\newblock {\em IEEE Transactions on Pattern Analysis and Machine Intelligence
  (TPAMI)}, 41(9):2251--2265, 2018.

\bibitem{fclswgan}
Yongqin Xian, Tobias Lorenz, Bernt Schiele, and Zeynep Akata.
\newblock {Feature generating networks for zero shot learning.}
\newblock In {\em IEEE Conference on Computer Vision and Pattern Recognition
  (CVPR)}, pages 5542--5551, 2018.

\bibitem{sdc}
Lu Yu, Bartlomiej Twardowski, Xialei Liu, Luis Herranz, Kai Wang, Yongmei
  Cheng, and Shangling Jui.
\newblock {Semantic drift compensation for class-incremental learning..}
\newblock In {\em IEEE Conference on Computer Vision and Pattern Recognition
  (CVPR)}, pages 6982--6991, 2020.

\bibitem{deepembed}
Li Zhang, Tao Xiang, and Shaogang Gong.
\newblock {Learning a deep embedding model for zero-shot learning.}
\newblock In {\em IEEE Conference on Computer Vision and Pattern Recognition
  (CVPR)}, pages 2021--2030, 2017.

\bibitem{zhang2021deep}
Yifan Zhang, Bingyi Kang, Bryan Hooi, Shuicheng Yan, and Jiashi Feng.
\newblock Deep long-tailed learning: A survey, 2021.

\end{thebibliography}
}

\clearpage
\appendix
\section*{\centering Appendix: Unseen Classes at a Later Time? No Problem}

This appendix provides additional details that could not be included in the main manuscript owing to space constraints. In particular, we provide: 
\vspace{-8pt}
\begin{itemize}
\setlength\itemsep{-0.05em}
    \item  {Details of datasets chosen for studies.}
    \item {Details of task-wise data splits for each of the three problem settings: Static, Dynamic, Online}
    \item{Metrics used for evaluating our model.}
    \item{Qualitative results depicting the similarity scores of visual features across tasks. }
    \item{Analysis of the generative module}
        \begin{itemize}
        \setlength\itemsep{-0.05em}
            \item {t-SNE visualizations to show contribution of various components of our approach and comparison with sequentially trained generative ZSL baseline.  }
            \item{Performance of the model on varying the number of replayed samples.}
        \end{itemize}
    \item{Implementation details of our method.}

\end{itemize}
Code for all experiments can be accessed at :  \href{https://github.com/sumitramalagi/Unseen-classes-at-a-later-time}{https://github.com/sumitramalagi/Unseen-classes-at-a-later-time}

\nocite{shivam,agem,apy,devise,zerovaegan,iisc,iisc2,azsl,ghosh,ewc,awa,zeroshotvis,lisgan,lwf,genclassinc,openlongtailrecognition,packnet,clbayes,catasinfer,cvae,sun,icarl,simplezsl,pnn,cada,cln,lsrgan,cub,bookworm,lzsl,zslcompre,fclswgan,sdc,deepembed,zhang2021deep}

\section{Dataset Details}
In this section, we provide a detailed description of the benchmark datasets used for evaluating our model in the \emph{static}, \emph{dynamic} and \emph{online} CGZSL settings. Following the existing literature \cite{cln,iisc,iisc2,azsl,ghosh} we assess our proposed model on five widely used benchmark datasets i.e  AWA1, AWA2, CUB, SUN, aPY, which are traditionally used for zero-shot learning. (Zero-shot recognition datasets are used since we require access to semantic attributes for addressing the CGZSL problem \cite{cln,iisc,iisc2,azsl,ghosh})\\
The Animals with Attributes dataset (AWA1 and AWA2) \cite{awa} consists of 50 classes of animals captured in diverse backgrounds. AWA1 consists of 30,475 images and AWA2 consists of 37,322 images. They are split into 40 seen classes and 10 unseen classes. The dataset also contains an 85-dimensional attribute vector for each class which is annotated by a human. Caltech UCSD Birds 200 (CUB)  \cite{cub} dataset consists of 11,788 images of birds in total, each of  which belongs to one of the  200 classes. In a standard generalized zero-shot learning (GZSL) setup, 150 of these classes are treated as seen and 50 classes are unseen. Each class in CUB has nearly 60 samples. In the CUB dataset, each class is also provided with a 312-dimensional human-annotated class attribute vector. The scene recognition dataset (SUN) \cite{sun} consists of 717 scenes or classes. Out of 717 classes, 645 classes are seen and the rest 72 are unseen. This dataset contains 14,340 fine-grained images and each class is associated with a 102-dimensional attribute. In aPY \cite{apy} dataset, there are 15,339 total images belonging to 32 classes. 20 of these classes are treated as seen and 12 are unseen. Each class is associated with a 64-dimensional attribute. We summarize the details of all datasets in Table \ref{table1_s}.

Following protocol in \cite{ghosh,iisc,iisc2,fclswgan,lsrgan}, the visual features for all datasets are extracted using  ResNet-101 pretrained on Imagenet dataset. We use the publicly available version of benchmark datasets provided by \cite{fclswgan}. 

\begin{table}
\resizebox{8cm}{!}{
\begin{tabular}{llllll}
\hline
Dataset & Attribute & \# Images & Seen Class & Unseen Class \\
\hline
AWA1    & 85        & 30,475     & 40         & 10           \\
AWA2    & 85        & 37,322     & 40         & 10           \\
aPY     & 64        & 15,339     & 20         & 12           \\
CUB     & 312        & 11,788     & 150         & 50       \\
SUN     & 102        & 14,340     & 645         & 72 \\  \bottomrule 
\end{tabular}}
\caption{Details of datasets used for zero-shot learning}
\label{table1_s}
\end{table}

\begin{table*}
\centering
\resizebox{17cm}{!}{%
  \begin{tabular}{{lllllllllllllllll}}
    \toprule
      &
      \multicolumn{2}{c}{Task-1} &
      \multicolumn{2}{c}{Task-2} &
      \multicolumn{2}{c}{Task-3} &
      \multicolumn{2}{c}{\textbf{Task-4 and more}} \\
      & {Seen (Replayed+New)} & {Unseen} & {Seen(Replayed+New)} & {Unseen} & {Seen(Replayed+New)} & {Unseen} & \textbf{...} & \textbf{...} \\
      \midrule
    \multicolumn{7}{c}{\textbf{aPY}}\\
    \hline
    Static&\hspace{10mm}{0+8}&24&\hspace{10mm}8+8&\hspace{2mm}16&\hspace{10mm}16+8&\hspace{2mm}8&\textbf{...}&\textbf{...}\\
    Dynamic&\hspace{10mm}0+5&\hspace{2mm}3&\hspace{10mm}5+5&\hspace{2mm}6&\hspace{10mm}10+5&\hspace{2mm}9&\textbf{...}&\textbf{...}\\
    Online&\hspace{10mm}0+4&\hspace{2mm}4&\hspace{10mm}4+5&\hspace{2mm}7&\hspace{10mm}9+5&\hspace{2mm}10&\textbf{...}&\textbf{...}\\
      \midrule
    \multicolumn{7}{c}{\textbf{AWA-1}}\\
    \hline
    Static&\hspace{10mm}0+10&\hspace{2mm}40&\hspace{10mm}10+10&\hspace{2mm}30&\hspace{10mm}20+10&\hspace{2mm}20&\textbf{...}&\textbf{...}\\
    Dynamic&\hspace{10mm}0+8&\hspace{2mm}2&\hspace{10mm}8+8&\hspace{2mm}4&\hspace{10mm}16+8&\hspace{2mm}6&\textbf{...}&\textbf{...}\\
    Online&\hspace{10mm}0+7&\hspace{2mm}3&\hspace{10mm}7+8&\hspace{2mm}5&\hspace{10mm}15+8&\hspace{2mm}7&\textbf{...}&\textbf{...}\\
      \midrule
    \multicolumn{7}{c}{\textbf{AWA-2}}\\
    \hline
    Static&\hspace{10mm}0+10&\hspace{2mm}40&\hspace{10mm}10+10&\hspace{2mm}30&\hspace{10mm}20+10&\hspace{2mm}20&\textbf{...}&\textbf{...}\\
    Dynamic&\hspace{10mm}0+8&\hspace{2mm}2&\hspace{10mm}8+8&\hspace{2mm}4&\hspace{10mm}16+8&\hspace{2mm}6&\textbf{...}&\textbf{...}\\
    Online&\hspace{10mm}0+7&\hspace{2mm}3&\hspace{10mm}7+8&\hspace{2mm}5&\hspace{10mm}15+8&\hspace{2mm}7&\textbf{...}&\textbf{...}\\
        \midrule
    \multicolumn{7}{c}{\textbf{CUB}}\\
    \hline
    Static&\hspace{10mm}0+10&\hspace{2mm}190&\hspace{10mm}10+10&\hspace{2mm}180&\hspace{10mm}20+10&\hspace{2mm}170&\textbf{...}&\textbf{...}\\
    Dynamic&\hspace{10mm}0+7&\hspace{2mm}2&\hspace{10mm}7+7&\hspace{2mm}4&\hspace{10mm}14+7&\hspace{2mm}6&\textbf{...}&\textbf{...}\\
    Online&\hspace{10mm}0+6&\hspace{2mm}3&\hspace{10mm}6+7&\hspace{2mm}5&\hspace{10mm}13+7&\hspace{2mm}7&\textbf{...}&\textbf{...}\\
        \midrule
    \multicolumn{7}{c}{\textbf{SUN}}\\
    \hline
    Static&\hspace{10mm}0+47&\hspace{2mm}670&\hspace{10mm}47+47&\hspace{2mm}623&\hspace{10mm}94+47&\hspace{2mm}576&\textbf{...}&\textbf{...}\\
    Dynamic&\hspace{10mm}0+43&\hspace{2mm}4&\hspace{10mm}43+43&\hspace{2mm}8&\hspace{10mm}86+43&\hspace{2mm}12&\textbf{...}&\textbf{...}\\
    Online&\hspace{10mm}0+42&\hspace{2mm}5&\hspace{10mm}42+43&\hspace{2mm}9&\hspace{10mm}85+43&\hspace{2mm}13&\textbf{...}&\textbf{...}\\
    
    \bottomrule
  \end{tabular}
  }
  \vspace{-4pt}
  \caption{Data-splits across all datasets for Static, Dynamic and Online settings. During each task, seen classes is the combination of replayed classes from previous task and newly added seen classes.}
  \vspace{-10pt}
\label{table2_s}
\end{table*}

\section{Task-wise Data Splits}
Unlike traditional GZSL methods where all classes are available during training/testing, continual GZSL (CGZSL) settings work on incremental tasks. As described in Sec. 3 of the main manuscript, the pattern in which new classes arrive in CGZSL depends on the setting (static, dynamic, online). Details of the task-wise split for standard zero-shot learning datasets with respect to various settings is described below:

\noindent \paragraph{Static CGZSL:} For the static CGZSL setting, we follow the dataset split mentioned in \cite{ghosh,azsl,iisc}. For a given task $T_t$, the first $t$ subsets i.e data belonging to the current and previous tasks are considered as seen while the rest are unseen. We divide AWA1 and AWA2 datasets into 5 tasks. The first task consists of 10 seen classes and 40 unseen classes. In each subsequent task we convert 10 of the unseen classes to seen. At the end of fifth task all the 50 classes of AWA1 and AWA2 dataset are converted to seen. SUN dataset is divided into 15 tasks with 47 unseen classes getting converted to seen in each task. CUB dataset is divided into 20 tasks where we incrementally convert 10 unseen classes into seen. aPY is split into 4 tasks, with each new task 8 previously unseen classes are converted to seen class.

\noindent \paragraph{Dynamic CGZSL:} In the dynamic CGZSL setting, new seen and unseen classes are added in each task. AWA1 and AWA2 datasets are divided into five tasks. In each task, 8 new seen and 2 new unseen classes are added. SUN dataset is divided into 15 task, where 43 seen classes and 4 unseen classes are added in each task. CUB dataset is divided into 20 tasks, where 7 seen classes and 2 unseen classes are added in each task. The aPY dataset consists of four tasks, with 5 seen classes and 3 unseen classes in each task.

\noindent \paragraph{Online CGZSL:} In our proposed online-CGZSL setting, each task has a disjoint set of seen and unseen classes. In addition, some of the previously unseen classes can turn into seen if the corresponding visual features become available for training in future tasks. To evaluate our model we consider the case where one of the previously unseen class is converted to seen class. AWA1 and AWA2 datasets are divided into five tasks. Every task consists of seven seen classes and three unseen classes. In addition, for each of task numbers two to five, one of the previously unseen classes is converted to a seen class. SUN dataset is divided into 15 tasks with 42 seen and 5 unseen classes. 6 seen and 3 unseen classes are added in each task for the CUB dataset over 20 tasks. aPY dataset is divided into four tasks with four seen and unseen classes. Similar to AWA1 and AWA2, one of the previously unseen classes is converted to a seen class during each task for CUB, SUN and aPY datasets. The data-splits for all settings are listed in the Table \ref{table2_s}.

\begin{figure*}[ht]
    \centering
    \includegraphics[width=0.9\textwidth]{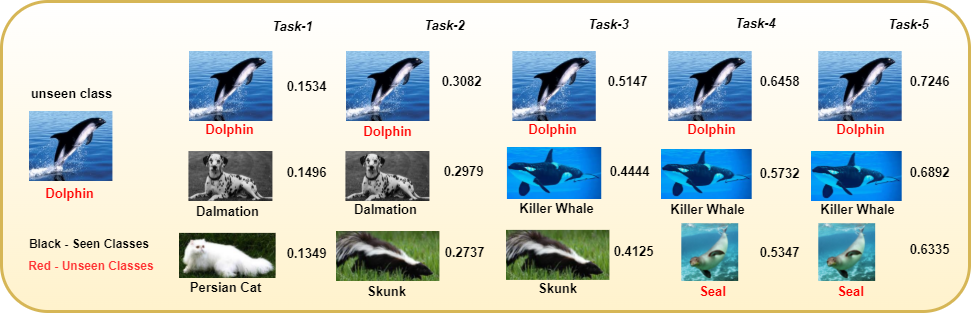}
    \caption{Cosine similarity scores of unseen class '\emph{dolphin}' of AWA2 dataset  w.r.t identifier projection. Top 3 cosine similarity scores shared with '\emph{dolphin}'  at every task is shown. Unseen classes are depicted in red, seen classes are in black.}
    \label{figure1}
\end{figure*}
\section{Evaluation Metrics}

\noindent \textbf{Static setting:} We follow the evaluation metrics mentioned in \cite{cln}: \newline
\begin{itemize}
\item{Mean Seen Accuracy (mSA) \begin{equation} \frac{1}{T}\sum _{t=1}^{T}CAcc({D}_{te}^{\le t},{A}^{\le t})\end{equation} }
\item{Mean Unseen Accuracy(mUA) \begin{equation} \frac{1}{T-1}\sum_{t=1}^{T-1}CAcc({D}_{te}^{>t},{A}^{>t})\end{equation}}
\item{ Mean Harmonic Accuracy (mH) \begin{equation} \frac{1}{T-1}\sum _{t=1}^{T-1}H({D}_{te}^{\le t},{D}_{te}^{>t},A)\end{equation} \newline}
\end{itemize}
\vspace{-1cm}
Here, $CAcc$ stands for per class accuracy, $H$ represents harmonic mean, $T$ denotes the total number of tasks, ${D}_{te}^{\le t}$ denotes test data till $t^{th}$ task,which according to setting-1, corresponds to seen data and ${D}_{te}^{>t}$ denotes test data of future tasks with respect to $t^{th}$ task. As per static setting, future task data is the unseen data. $A$ denotes the set of all attributes.

\vspace{8pt}
\noindent \textbf{Dynamic setting:} We use a similar evaluation metrics as mentioned in \cite{iisc,iisc2} :
\begin{itemize}
\item{Mean Seen Accuracy (mSA) \begin{equation} \frac{1}{T}\sum _{t=1}^{T}CAcc({D}_{{te}_{s}}^{\le t},{A}^{\le t})\end{equation} }
\item{Mean Unseen Accuracy(mUA) \begin{equation} \frac{1}{T}\sum _{t=1}^{T}CAcc({D}_{{te}_{u}}^{\le t},{A}^{\le t})\end{equation}}
\item{ Mean Harmonic Accuracy (mH) \begin{equation} \frac{1}{T}\sum _{t=1}^{T}H({D}_{{te}_{s}}^{\le t},{D}_{{te}_{u}}^{\le t},{A}^{\le t})\end{equation} \newline}
\end{itemize}

Here, $CAcc$ stands for per class accuracy, $H$ represents harmonic mean, $T$ denotes the total number of tasks, ${D}_{{te}_{s}}^{\le t}$ denotes test data of seen classes till $t^{th}$  and ${D}_{{te}_{u}}^{\le t}$ denotes test data of unseen classes till $t^{th}$ task. ${A}^{\le t}$ denotes the set of all attributes encountered so far.

\vspace{8pt}
\noindent \textbf{Online setting:} We use the evaluation metrics proposed in dynamic setting considering the updated set of seen and unseen classes for calculating accuracy.

We re-calculate each task's accuracy in order to obtain the average accuracy for a given task.

\subsection{Additional Evaluation Metrics}
\noindent \textbf{mAUSUC:}  \cite{cln} adopted the mean area under seen/unseen curve (mAUSUC) as metric for measuring the performance of CGZSL models across all tasks.  
mAUSUC is given by:
\begin{equation}mAUSUC(F)=\frac{1}{T}\sum _{t=1}^{T}AUSUC(F,{D}_{te}^{\le t},{A}^{\le t})\end{equation}
\noindent where $T$ is the total number of tasks encountered so far, ${D}_{te}^{\le t}$ is the test data consisting of both seen and unseen classes and ${A}^{\le t}$ is the set of attributes encountered so far. 
We compare our mAUSUC score with recent state-of-the-art approaches such as  NM-ZSL \cite{cln}, Tf-GCZSL \cite{iisc2}
and A-CZSL \cite{azsl}. Figure \ref{figure7} shows task-wise mAUSUC on AWA2 dataset in all three settings. A higher value of mAUSUC indicates that the model is better able to handle bias between seen and unseen classes. We observe that our model's performance is superior to the existing state-of the art CGZSL methods.

\begin{figure*}[ht]
    \centering
    \includegraphics[width=1\textwidth]{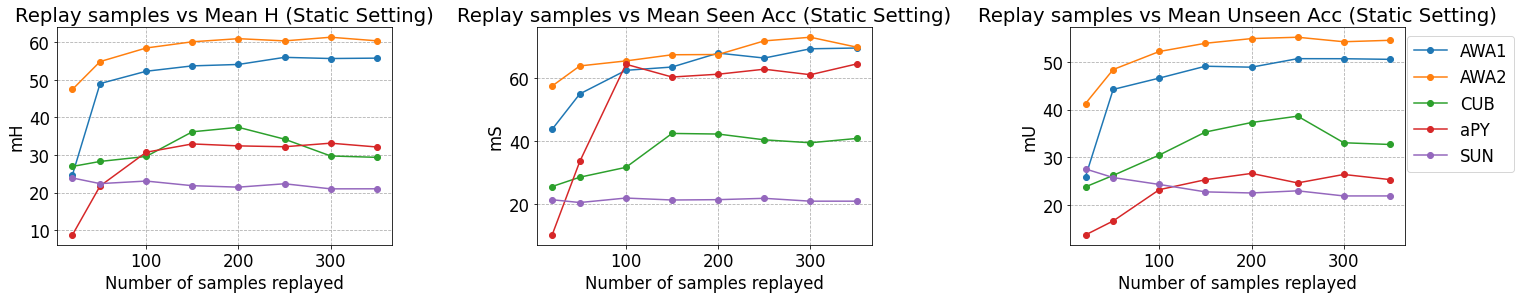}
    \caption{Number of replayed samples vs Mean Harmonic accuracy (mH), Mean Seen accuracy (mS) and Mean Unseen accuracy (mU) on all datasets in Static-CGZSL setting. We observe that fine-grained datasets like CUB and SUN perform well when the number of replayed samples is less. Performance of coarse-grained datasets like AWA1 and AWA2, saturates when the number of replayed samples is more than two hundred. }
    \label{figure4}
\end{figure*}
\section{Task-wise Similarity Scores of Visual Features}

 In the CGZSL settings, seen/unseen classes are added incrementally. Our model retains the previously learned knowledge using incremental bi-directional alignment (Sec. 4.3) and generative replay of visual features from the previous classes (Sec. 4.4). The incremental bi-directional alignment loss is a combination of nuclear loss and semantic alignment loss. Nuclear loss helps in aligning identifier projections in accordance with the real visual features and semantic alignment loss aids in strengthening semantic relationships and knowledge transfer as the visual space evolves and new classes are added over time.

In this section, we analyze how applying  semantic alignment mechanism incrementally helps in generating better visual features by strengthening semantic relationships as new classes are added. 
Since real visual features for unseen classes are not available during training, the semantic alignment loss $L_{sal}$, tries to leverage semantic structure of all the classes encountered so far and generate accurate unseen features. The seen normalized loss $L_{snl}$ discussed in Sec. 4.2, helps the discriminator to place the identifier projection of unseen classes close to the generated unseen features. Furthermore, the semantic alignment loss is applied with respect to $n_{c}$ nearest neighbours of class $c$. As the class distribution changes with time due to dynamic addition of new classes, the visual space evolves and the nearest neighbours for a particular unseen class changes. As new classes that are semantically similar to a given unseen class arrive, performing semantic alignment with nearest classes incrementally helps in further enhancing semantic relationships w.r.t new class distribution and improving the quality of generated unseen class features.

\begin{figure*}[ht]
    \centering
    \includegraphics[width=1\textwidth]{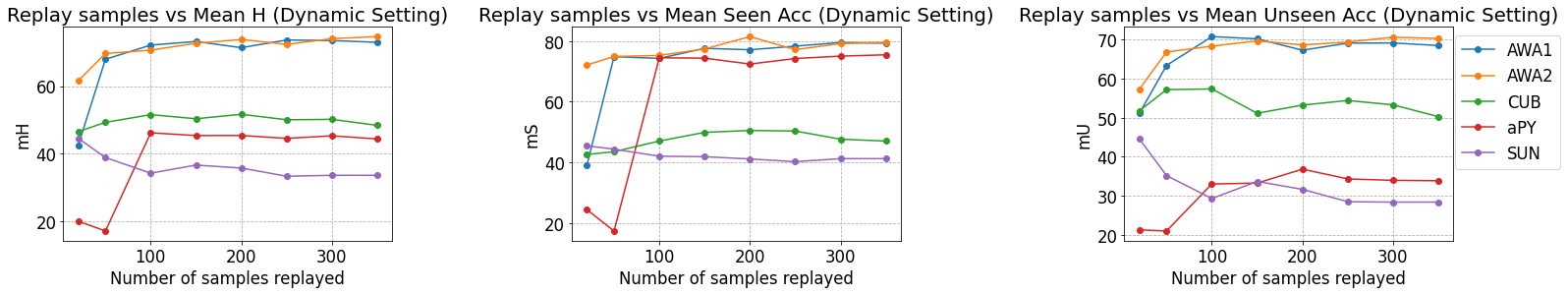}
    \caption{\footnotesize Number of replayed samples vs Mean Harmonic accuracy (mH), Mean Seen accuracy (mS) and Mean Unseen accuracy (mU) on all datasets in Dynamic-CGZSL setting. We observe that performance of coarse-grained datasets like AWA1 and AWA2 increases, if the number of replayed samples is more than hundred.}
    \label{figure5}
\end{figure*}
\begin{figure*}[ht]
    \centering
    \includegraphics[width=1\textwidth]{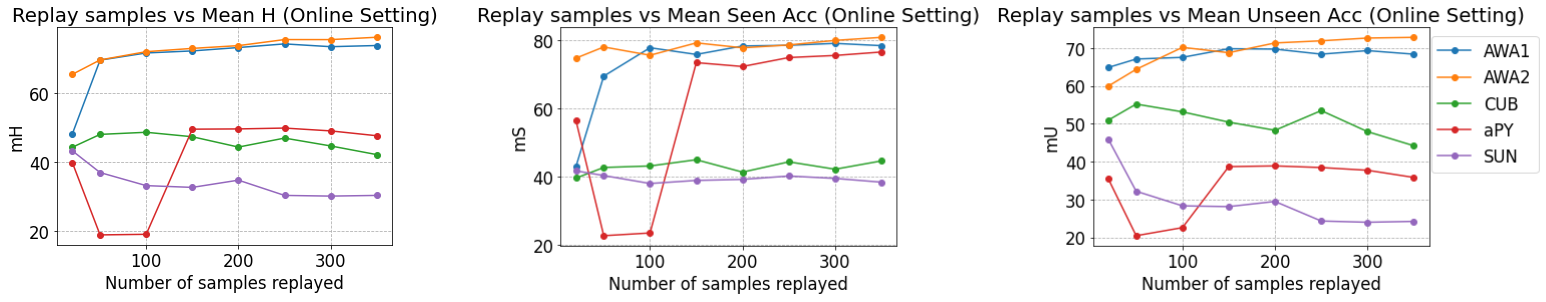}
    \caption{\footnotesize Number of replayed samples vs Mean Harmonic accuracy (mH), Mean Seen accuracy (mS) and Mean Unseen accuracy (mU) on all datasets in Online-CGZSL setting. For fine-grained datasets like SUN, we can notice that replaying lesser samples helps to boost the performance as the number of samples per class are less in SUN dataset. Coarse-grained datasets like AWA1, AWA2 and aPY perform well when the number of replayed samples are more.}
    \label{figure6}
\end{figure*}

Figure \ref{figure1} shows cosine similarities calculated during inference between the unseen class '\emph{dolphin}' of the AWA2 dataset and the most closely related seen-unseen classes (in terms of identifier projections) across different tasks. We observe that the alignment of the identifier projections with semantically similar classes during each task. While the model predicts '\emph{dolphin}' correctly across the 5 tasks, during task-1, with the available seen and unseen classes, '\emph{dolphin}' shares a cosine similarity score of 0.1349 with the identifier projection of the unseen class '\emph{persian cat}'. With the addition of new seen and unseen classes during task-2, '\emph{dolphin}' now shares a cosine similarity of 0.2737 with '\emph{skunk}' rather than '\emph{persian cat}' whose cosine similarity dropped to 0.1024 . This distinctly shows that the semantic alignment loss is helping to improve the representation of generated unseen features. The generated unseen features guide discriminator in mapping identifier projections, which in turn aid in classification. It can be seen that at the end of task-5, '\emph{dolphin}' shares high cosine similarity scores with identifier projection of classes '\emph{killer whale}' and '\emph{seal}' which are visually close. This shows how the model leverages current semantic structure to learn better identifier projections as new classes are added.

\begin{figure*}[ht]
    \centering
    \includegraphics[width=0.8\textwidth]{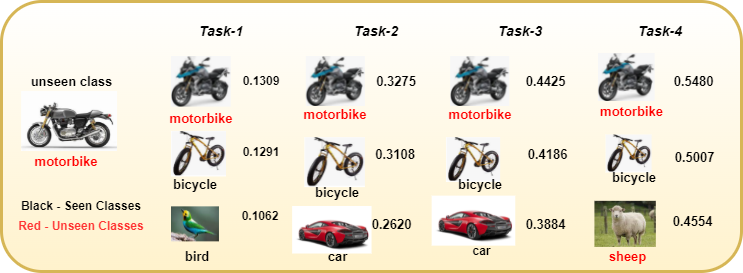}
    \caption{Cosine similarity scores of unseen class '\emph{motorbike}' of aPY dataset. Top three cosine similarity scores shared with '\emph{motorbike}' at every task is shown. Unseen classes are depicted in red, seen classes are in black. }
    \label{figure2}
\end{figure*}
\begin{figure*}[ht]
    \centering
    \includegraphics[width=0.9\textwidth]{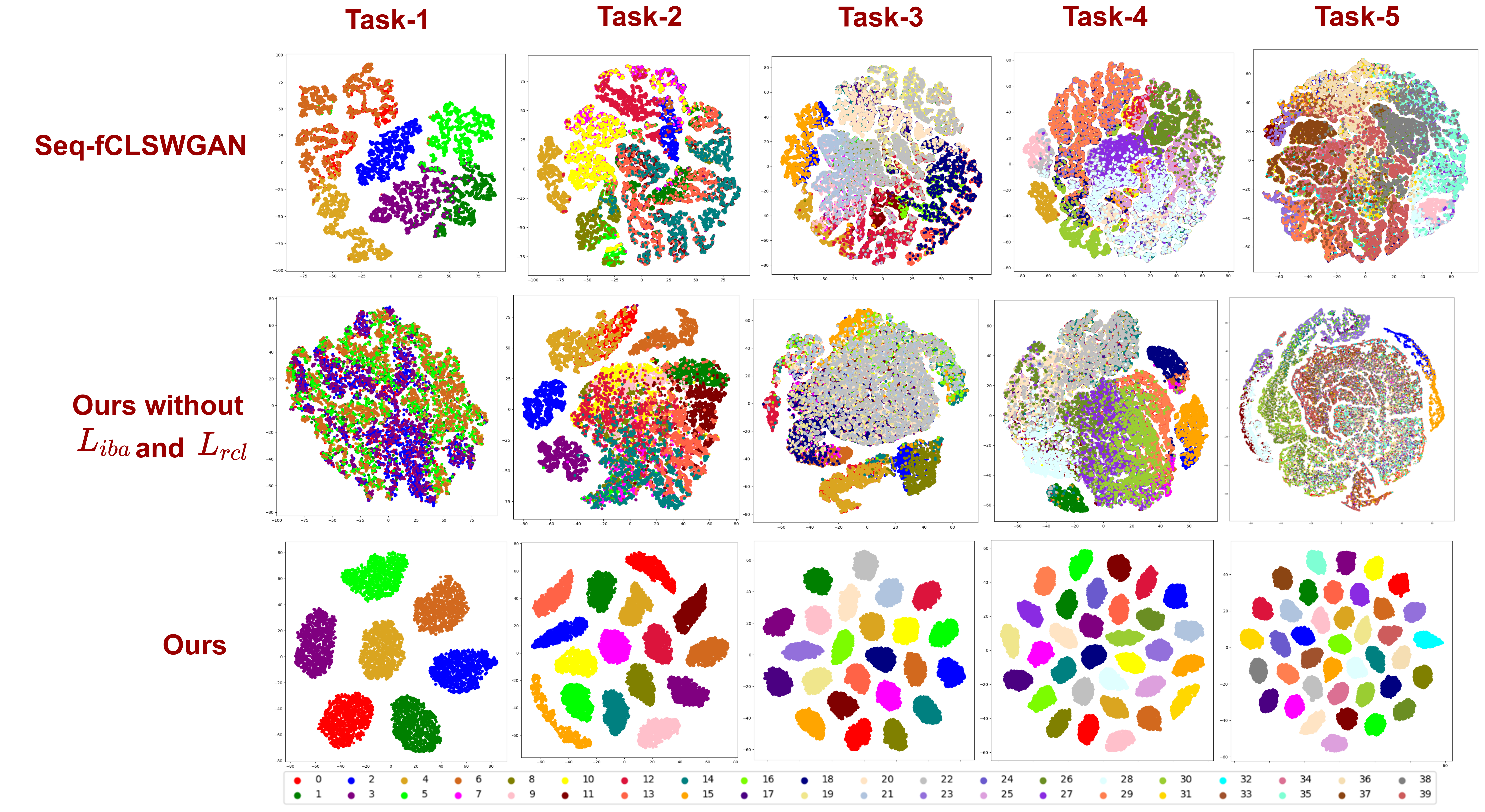}
    \caption{t-SNE visualizations of visual features generated by Seq-fCLSWGAN (Row 1), our method without incremental bi-directional alignment ($L_{iba}$) and real classification loss ($L_{rcl}$) (Row 2) and our overall approach (Row 3) during various tasks for AWA1 dataset. Different colors depict different seen classes.}
    \label{figure3}
\end{figure*}

More such examples are shown in Figure 4 presented in the main manuscript (which we explain here owing to space constraints) and Figure \ref{figure2}. In Figure 4, for the unseen class '\emph{sheep}' of the AWA2 dataset encountered during task-1, the similarities with identifier projection of '\emph{killer whale}' and '\emph{skunk}' classes are 0.2765 and 0.2533. With the addition of new seen and unseen classes during task-2, '\emph{sheep}' now shares a cosine similarity of 0.4380 with '\emph{gorilla}' rather than '\emph{killer whale}' whose cosine similarity dropped to 0.2543 . Note however that the '\emph{sheep}' class is classified correctly in all tasks. Figure \ref{figure2} illustrates an example from the aPY dataset, where the unseen class '\emph{motorbike}' shares high cosine similarity score with the identifier projection of bicycle class. Since '\emph{bicycle}' and '\emph{motorbike}' are visually very similar, '\emph{bicycle}' has the highest cosine similarity with '\emph{motorbike}' in spite of new classes being added. We can observe that cosine similarity between visually related samples keeps increasing as new tasks arrive, ensuring better alignment between the identifier projections.

\section{Analysis of Performance of Generative Replay Module}
\noindent \textbf{t-SNE plot visualization:} Our model uses generative replay to overcome catastrophic forgetting. We visualize the generated visual features of AWA1 dataset per task (Figure \ref{figure3}) and observe that generated visual features form well-defined clusters. The features belonging to same class are grouped together and far from other classes, this signifies that the generator is able to generate discriminative visual features. Since f-CLSWGAN \cite{fclswgan} is a GAN-based approach for solving GZSL problems, we compare the visual features generated by our model with the sequential version of f-CLSWGAN (Seq-fCLSWGAN) \cite{fclswgan}. We notice that well-defined clusters are formed during task-1 of Seq-fCLSWGAN training, but subsequently it tends to forget the acquired knowledge and the newly generated features tend to get mixed up in the visual space. We perform ablation study on the proposed approach and show that removing incremental bi-directional alignment and classification loss deforms the clusters across all tasks. 
\begin{figure*}[ht]
    \centering
    \includegraphics[width=1\textwidth]{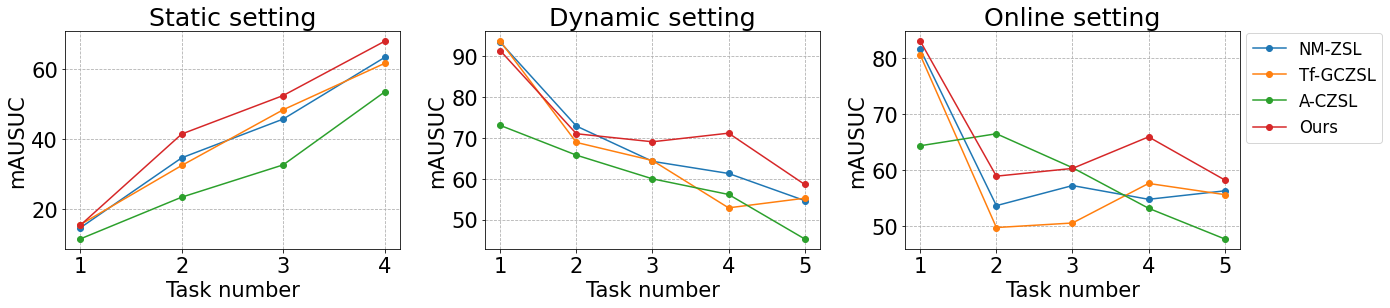}
    \caption{Task-wise mean AUSUC values for static (left), dynamic (center) and online (right). }
    \label{figure7}
\end{figure*}

\vspace{8pt}
\noindent \textbf{Varying number of replayed samples:} We evaluated the performance of our model by varying the number of samples replayed. We compare the number of samples with mean harmonic accuracy (mH), mean seen accuracy (mS) and mean unseen accuracy (mU) in all the three settings and all the datasets. We plot the results in Figure \ref{figure4} (static) , \ref{figure5} (dynamic) and \ref{figure6} (online). We observe that across all settings AWA1, AWA2 perform well when the number of replayed samples is more than 100. We use a replay of 300 for AWA1, AWA2 dataset in our main manuscript. aPY dataset has only 32 classes in total, hence to avoid overfitting we replay only 150 samples per class.

SUN and CUB are fine-grained datasets, and each class has limited data which makes these datasets challenging. In order to mimic the original dataset, we replay only 150 samples per task for CUB and only 20 samples per task for the SUN dataset.

\section{Implementation Details}
Our model is implemented using Pytorch-1.4.0 and CUDA-11.2. 

The proposed approach consists of a generator $G$ and discriminator $D$.

We use Adam optimizer with a learning rate of 0.005 and weight\_decay of 0.00001 for all settings and datasets. We normalize both target image and attributes before calculating cosine similarity.
The total G loss is given by: $L_{G}^{t}$ = $\lambda_{1}$ $L_{GAN}$ + $\lambda_{2}$ $L_{pcl}$ +  $\lambda_{4}$ $L_{iba}$ and
the overall D loss is given by: $L_{D}^{t}$ = $\lambda_{1}$ $L_{GAN}$ + $\lambda_{2}$ $L_{rcl}$ + $\lambda_{3}$ $L_{snl}$, where $\lambda_{1}$, $\lambda_{2}$, $\lambda_{3}$, $\lambda_{4}$ are 1.

\end{document}